\documentclass[11pt]{article}

\usepackage[final]{acl}
\usepackage{times}
\usepackage{latexsym}
\usepackage[T1]{fontenc}
\usepackage[utf8]{inputenc}

\usepackage{microtype}

\usepackage{inconsolata}

\usepackage{graphicx}

\usepackage[textsize=tiny]{todonotes}

\usepackage[most]{tcolorbox}
\usepackage{comment}
\usepackage[record,abbreviations]{glossaries-extra}
\glsdisablehyper
\newabbreviation{llm}{LLM}{Large Language Model}
\newabbreviation{nl}{NL}{Natural Language}
\newabbreviation{pl}{PL}{Formal Language} % Context: "formal languages (PL)"
\newabbreviation{atp}{ATP}{Automated Theorem Proving}

\newabbreviation{em}{EM}{Exact Match}
\newabbreviation{bleu}{BLEU}{Bilingual Evaluation Understudy}

\newabbreviation{rag}{RAG}{Retrieval-Augmented Generation}
\newabbreviation{rl}{RL}{Reinforcement Learning}

\newabbreviation{cot}{CoT}{Chain-of-Thought}

\newabbreviation{ea}{TA}{Testing Accuracy}
\usepackage{amsmath}
\usepackage{pifont}
\usepackage{tabularx}
\usepackage{caption, booktabs}
\usepackage{algorithm}
\usepackage{amssymb}
\usepackage{setspace}
\usepackage{bbm}
\usepackage{arydshln}
\usepackage{makecell}

\usepackage{todonotes}

\usepackage{stmaryrd}

\makeatletter
\newcommand\footnoteref[1]{\protected@xdef\@thefnmark{\ref{#1}}\@footnotemark}
\makeatother
\newcolumntype{P}[1]{>{\centering\arraybackslash}p{#1}}

\usepackage{xspace}

\newcommand{\ours}[0]{\textsc{T$^2$}\xspace}

\newtcbox{\codebox}{on line, boxrule=0pt, boxsep=0pt, colback=gray!10, colframe=gray!10, arc=2pt, left=2pt, right=2pt, top=1pt, bottom=1pt}
\newcommand{\code}[1]{\codebox{\texttt{#1}}}

\usepackage{graphicx}
\usepackage{adjustbox}
\makeatletter
\newcommand{\thickhline}{
\noalign {\ifnum 0=`}\fi \hrule height 1pt
    \futurelet \reserved@a \@xhline
}
\newcolumntype{"}{@{\hskip\tabcolsep\vrule width 1pt\hskip\tabcolsep}}
\makeatother
\usepackage{mathabx}
\usepackage{amsfonts}

\makeatletter
\newcommand*{\blackleq}{
    \mathrel{
        \mathpalette\@blackleq{}
    }
}
\newcommand*{\@blackleq}[2]{
    \vcenter{
        \m@th
        \setbox0=\hbox{$#1\mkern3mu$}
        \setbox2=\hbox{$#1\vcenter{}$}
        \setbox4=\hbox{\raisebox{-\ht2}[.2pt][.2pt]{$#1-$}}
        \hbox{$#1\blacktriangleleft$}
        \nointerlineskip
        \kern\wd0
        \copy4
    }
}
\makeatother
\usepackage{dsfont}
\usepackage{multirow}
\usepackage{kotex} % Menu > TeX live version: 2021 is required

\usepackage{mathtools}

\definecolor{my_blue}{RGB}{0,112,192}

\DeclareRobustCommand{\xmark}{\textcolor{red}{\ding{55}}}
\DeclareRobustCommand{\cmark}{\textcolor{green!60!black}{\ding{51}}}

\usepackage{placeins}

\usepackage{tikz}

\usepackage{minted}
\usepackage{fvextra}

\DefineVerbatimEnvironment{WideMinted}{Verbatim}
{breaklines, fontsize=\small, frame=lines, linenos, breakanywhere, xleftmargin=0pt, xrightmargin=0pt, width=\textwidth}

\usepackage{stackengine}

\makeatletter
\def\th@definition{%
    \normalfont % Body font
    \thm@headpunct{.}% Add a period after the theorem number
}
\makeatother

\usepackage{enumitem}
\usepackage{wrapfig}
\usepackage{subcaption}

\usepackage{listings}

\definecolor{codegray}{rgb}{0.5,0.5,0.5}
\definecolor{codepurple}{rgb}{0.58,0,0.82}
\definecolor{backcolour}{rgb}{0.95,0.95,0.92}

\lstdefinelanguage{Lean}{
keywords={def, theorem, lemma, example, where, match, with, end, class, instance, structure, inductive},
keywordstyle=\color{blue}\bfseries,
ndkeywords={Type, Prop, Nat},
ndkeywordstyle=\color{teal}\bfseries,
identifierstyle=\color{black},
sensitive=true,
comment=[l]{--},
morecomment=[s]{/-}{-/},
commentstyle=\color{codegray}\ttfamily,
stringstyle=\color{red}\ttfamily,
morestring=[b]",
basicstyle=\ttfamily\footnotesize,
breaklines=true,
keepspaces=true,
showstringspaces=false,
frame=none,
backgroundcolor=\color{backcolour},
numbers=none,
literate={{<}}{{$\langle$}}1 {{>}}{{$\rangle$}}1 % prettify brackets if needed, or remove
}

\lstdefinestyle{promptstyle}{
    backgroundcolor=\color{backcolour},
    commentstyle=\color{codegray},
    keywordstyle=\color{magenta},
    numberstyle=\tiny\color{codegray},
    stringstyle=\color{codepurple},
    basicstyle=\ttfamily\footnotesize,
    breakatwhitespace=false,
    breaklines=true,
    captionpos=b,
    keepspaces=true,
    numbers=none,
    numbersep=5pt,
    showspaces=false,
    showstringspaces=false,
    showtabs=false,
    tabsize=2
}

\tcbuselibrary{listings, breakable}

\newtcolorbox{systemprompt}{
    colback=red!5!white,
    colframe=red!75!black,
    title=System Prompt,
    fonttitle=\bfseries,
    boxrule=0.5mm,
    arc=2mm,
    left=2mm, right=2mm, top=2mm, bottom=2mm,
    breakable
}

\newtcolorbox{userprompt}{
    colback=blue!5!white,
    colframe=blue!75!black,
    title=User Prompt,
    fonttitle=\bfseries,
    boxrule=0.5mm,
    arc=2mm,
    left=2mm, right=2mm, top=2mm, bottom=2mm,
    breakable
}

\newtcolorbox{modelresponse}{
    colback=green!5!white,
    colframe=green!75!black,
    title=AI Response,
    fonttitle=\bfseries,
    boxrule=0.5mm,
    arc=2mm,
    left=2mm, right=2mm, top=2mm, bottom=2mm,
    breakable
}
\newtcblisting{leancode}{
    listing only,
    colback=white,
    colframe=gray!50,
    colbacktitle=gray!20,
    coltitle=black,
    title={\scriptsize\textbf{Lean 4}},
    fonttitle=\sffamily,
    boxrule=0.5pt,
    arc=1pt,
    left=1mm, right=1mm, top=1mm, bottom=1mm,
    listing options={
            language=Lean,
            style=promptstyle,
        }
}

\newtcblisting{textcode}{
    listing only,
    colback=white,
    colframe=gray!50,
    colbacktitle=gray!20,
    coltitle=black,
    title={\scriptsize\textbf{Natural Language}},
    fonttitle=\sffamily,
    boxrule=0.5pt,
    arc=1pt,
    left=1mm, right=1mm, top=1mm, bottom=1mm,
    listing options={
            style=promptstyle,
        }
}

\usepackage[hang, flushmargin]{footmisc}
\usepackage{hyperref}

\usepackage{cleveref}% Load AFTER hyperref!
\title{
Benchmarking Testing in Automated Theorem Proving 
}

\author{
    Jongyoon Kim\textsuperscript{\rm 1}\thanks{~~Both authors contributed equally to this research.} 
    Hojae Han\textsuperscript{\rm 2}\textsuperscript{*}
    Seung-won Hwang\textsuperscript{\rm 1}\thanks{~~Corresponding Author}~\\
    Interdisciplinary Program in Artificial Intelligence, Seoul National University\textsuperscript{\rm 1}\\
    Electronics and Telecommunications Research Institute\textsuperscript{\rm 2}\\
    \texttt{\{john.jongyoon.kim,seungwonh\}@snu.ac.kr} \\
    \texttt{\{hojae.han\}@etri.re.kr} \\
}

\begin{document}
\maketitle

\begin{abstract}
    Recent advances in large language models (LLMs) have shown promise in formal theorem proving, yet evaluating semantic correctness remains challenging. 
    Existing evaluations rely on indirect proxies such as lexical overlap with human-annotated proof,
    %NLI with back-translation, prover-based equivalence, 
    or expensive manual inspection.
    Inspired by the shift from lexical comparison to test-based evaluation in code generation, we propose \ours, a framework that evaluates the semantic correctness of formal theorems: a generated theorem is considered correct only if all dependent successor theorems compile successfully, analogous to integration testing.
    We construct a benchmark from 5 real-world Lean 4 repositories, comprising 2,206 problems paired with 41 successor theorems on average, automatically extracted without human effort.
    Experiments demonstrate that while state-of-the-art models achieve high compilation success, they perform significantly worse under our semantic metric.
    The best model, Claude-Sonnet-4.5, achieves only 38.9\% Testing Accuracy on the full set, given both natural language proof and successor theorems as context, revealing a critical gap in current theorem generation capabilities.\footnote{Implementation and dataset are available at: \url{https://github.com/ldilab/T2}.}

\end{abstract}

\section{Introduction}
\label{sec:intro}

\begin{figure}[t]
    \centering
    \includegraphics[width=\linewidth]{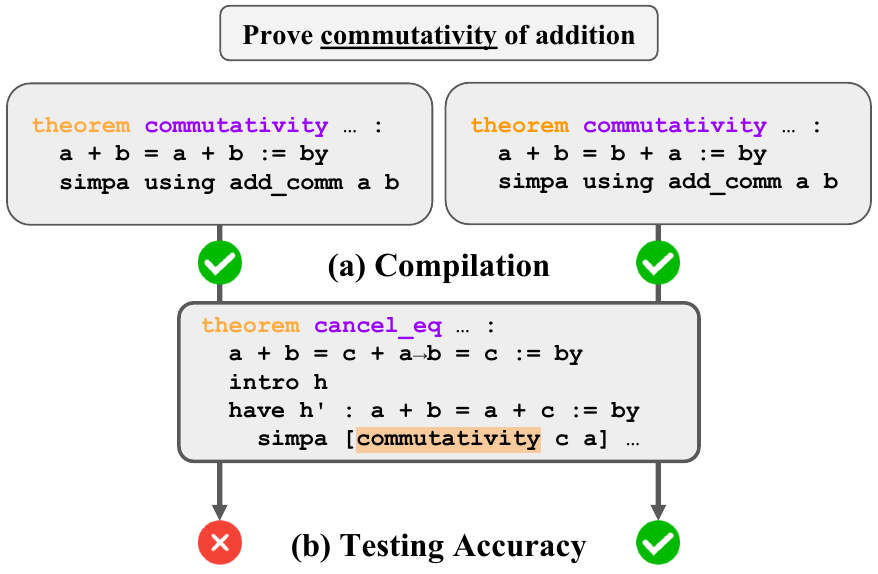}
    \caption{
    Two candidate theorems for proving commutativity of addition: one correct ($a+b=b+a$) and one tautology ($a+b=a+b$).
    (a) Under compilation-based evaluation, both candidates pass, as each is logically valid in isolation.
    (b) Under testing-based evaluation, the tautology is exposed: the successor theorem (\texttt{cancel\_eq}) fails when it depends on the incorrect candidate, while it compiles successfully with the correct one.
    }
    % \caption{
    %     Both (a) and (b) compile successfully under the existing evaluation. 
    %     However, (a) proves a tautology instead of the intended commutativity, a false positive that compilation alone cannot detect. 
    %     Our execution-based evaluation exposes this error: the successor theorem (\texttt{cancel\_eq}) fails when the incorrect theorem is substituted, while it compiles successfully with the correct one in (b).
    % }
    % (a)는 lexical-based로 해서 두 예제 다 pass시키고
    % (b) ours는 successor theorem으로 평가해서 a + b = a + b를 걸러내는거죠
    % \caption{
    % (a) Existing benchmarks evaluate a generated theorem, verifying only whether it compiles. 
    % Dependent theorems (dashed borders) are ignored, so semantic misalignment, such as proving a tautology instead of the intended property, goes undetected.
    % (b) In contrast, our approach integrates the generated theorem into a dependency graph and verifies whether every successor theorem successfully compiles, enabling \textbf{execution-based semantic evaluation}.
    % }
    \label{fig:introductory}
\end{figure}

\begin{table}[t]
  \centering
  \scriptsize
  \tabcolsep=1.5pt
  \scalebox{0.95}{
      \renewcommand{\arraystretch}{1.1}
      \begin{tabular}{l l c c c}
        \toprule
        \textbf{Benchmark} & \textbf{Eval. Method} & \textbf{Annot.-Free} & \textbf{Semantic} \\
        \midrule
        MiniF2F \cite{zheng2022minif2fcrosssystembenchmarkformal}                 & ---        & \xmark & \xmark \\
        FormalMATH \cite{yu2025formalmathbenchmarkingformalmathematical}           & ---     & \xmark & \xmark \\
        PutnamBench \cite{tsoukalas2024putnambenchevaluatingneuraltheoremprovers}  & ---    & \xmark & \xmark \\
        MiniF2F-v2 \cite{ospanov2025minif2fleanrevisitedreviewinglimitations}     & Human     & \xmark & \cmark \\
        ProofNet \cite{azerbayev2023proofnetautoformalizingformallyproving}        & Lexical       & \cmark & \xmark \\
        Lean-Workbook \cite{ying2025leanworkbooklargescalelean}                    & Lexical        & \cmark & \xmark \\
        Con-NF \cite{liu2025rethinking-connf}                                      & Prover        & \cmark & \xmark \\
        ProofNetVerif \cite{poiroux2025reliable-proofnetverif}                     & Prover       & \cmark & \xmark \\
        \midrule
        \textbf{\ours (Ours)}                                                      & \textbf{Testing} & \cmark & \cmark \\
        \bottomrule
      \end{tabular}
  }
  \caption{
    % Comparison of semantic correctness evaluation across formal theorem benchmarks.
    % All benchmarks verify logical validity via compilation. 
    % We compare methods for evaluating semantic correctness.
    % \textit{Annot.-Free}: operates without human evaluation.
    % \textit{Semantic}: verifies semantic correctness.
    % % $\triangle$ indicates an indirect proxy.
    Comparison of semantic correctness evaluation across formal theorem benchmarks. 
    All benchmarks verify logical validity via compilation. 
    \textbf{Annot.-Free}: operates without human annotation or evaluation. 
    \textbf{Semantic}: directly verifies semantic correctness.
}
  \label{tab:contribution}
\end{table}
With the rapid advancement of \glspl{llm}, there has been growing interest in applying these models to Automated Theorem Proving (ATP), where \gls{nl} problems are formalized as theorems in proof assistants such as Lean, with approaches ranging from fine-tuned provers~\cite{wang2025kiminaproverpreviewlargeformal, xin2024deepseekproveradvancingtheoremproving} to in-context learning agents 
\cite{thakur2024an, kumarappan2025leanagent, varambally2025hilbert}. 
Recent benchmarks demonstrate that \glspl{llm} can generate valid formal proofs spanning high school olympiad- to undergraduate-level mathematics~\cite{zheng2022minif2fcrosssystembenchmarkformal, azerbayev2023proofnetautoformalizingformallyproving, yu2025formalmathbenchmarkingformalmathematical}, where correctness is verified by checking whether the generated theorem compiles, confirming its syntactic and logical validity.

However, compilation alone does not guarantee \emph{semantic} correctness. 
As illustrated in \Cref{fig:introductory}-(a), an \gls{nl} specification may ask to prove commutativity ($a + b = b + a$), while the generated theorem is a tautology ($a + b = a + b$). 
As this tautology compiles, which only verifies its logical validity,
the intended meaning may be lost.
% one may fail to capture the intended meaning is lost.

Existing benchmarks have recognized this limitation and proposed various proxy metrics to assess semantic correctness, as summarized in \Cref{tab:contribution}. 
These approaches evaluate the generated theorem by comparing with a ground-truth proof reference, using BLEU~\cite{zheng2022minif2fcrosssystembenchmarkformal}, or equivalence checking using linguistic entailment such as natural language inference~\cite{ying2025leanworkbooklargescalelean}, or using a compiler built-in prover such as BEq~\cite{liu2025rethinking-connf}.
Alternatively, humans may investigate the generated theorem
without any reference annotation~\cite{ospanov2025minif2fleanrevisitedreviewinglimitations, tsoukalas2024putnambenchevaluatingneuraltheoremprovers, yu2025formalmathbenchmarkingformalmathematical}.
Desirably, this evaluation should be automatic, requiring neither human
evaluation nor reference annotation, while
directly verifying semantic correctness.

Towards this goal, we take inspiration from integrated testing in
software engineering, 
%Rather than evaluating the generated theorem itself, we take inspiration from integration testing in software engineering, 
where a component's correctness is verified, not in isolation, but by executing the modules that depend on it. 
Drawing on the \textit{Curry-Howard correspondence}~\cite{curry--howard_correspondence}, where proofs correspond to programs, we observe that a successor theorem that uses a generated lemma is analogous to a module that calls a function. 
That is, if the lemma is semantically incorrect, the successor theorem
%can no longer be constructed, just as a faulty function causes its callers to fail. 
cannot be correct.
Based on this insight, we propose evaluating a generated theorem by whether its successor theorems, those that invoke or build upon it, compile successfully, as illustrated in \Cref{fig:introductory}-(b). 
This approach is fully automatic, requires no external references, and directly verifies semantic correctness through successor compilation.

Building on this idea, we introduce \textbf{\gls{ea}}, the first testing-based metric for semantic evaluation of formal theorems, and \textbf{T}heorem \textbf{T}esting (\textbf{\ours}), a large-scale Lean benchmark designed to evaluate the generated theorem by testing it with successor theorems.
%it. 
\ours comprises 2,206 theorems curated from real-world Lean repositories, each paired with successor theorems that serve as integration test cases. 
We further identify a Hard subset of 389 problems that requires the model to generate both an auxiliary proposition and a proof of the target theorem.

Across experiments with 18 open and closed-source models, we observe a substantial gap between compilation and testing accuracy: models that appear successful under traditional measures often fail semantically. 
Our key findings are:
\begin{itemize}
    \item \textbf{High false positive rate of existing metrics}: Up to 93.1\% of compilable theorems are semantically incorrect, and BLEU fails to distinguish correct from incorrect outputs.
    \item \textbf{Overfits to compilation}: Specialized proving models overfit to syntax and score lower on semantic correctness than similarly sized general-purpose models.
    \item \textbf{Dependency structure enables 1k+ test cases}: \gls{ea} becomes more discriminative as successor coverage grows, with the majority of problems verified at depth 7 and an average of 1.6k successor theorems per problem at that depth.
    \item \textbf{Successor context improves generation}: Providing successor theorems as context consistently improves semantic correctness, while NL proofs alone do not.
\end{itemize}
% ==================================================================
% Outline: Related Work
% ==================================================================
% Structure: Following Professor's Guidance (Theoretical Foundations)
%
% 1. FTP Benchmarks
%    - Existing benchmarks (MiniF2F, ProofNet, etc.).
%    - Focus: Syntactic/Logical validity (Compilation) vs. Semantic Intent.
%    - Gap: Lack of behavioral correctness checks; reliance on proxies (BLEU, NLI).
%    - Distinction: We use behavioral evidence (downstream execution).
%
% 2. Eliminators under Curry--Howard
%    - Theory: Proofs as programs, Usage as Eliminators.
%    - Perspective: Observational Equivalence (do proofs behave same under eliminators?).
%    - Distinction: We treat downstream theorems as eliminators for generated theorems.
%
% 3. Differential Testing and Program Equivalence
%    - SW Engineering: Testing behavior vs. proving equivalence.
%    - Metric: Output agreement on shared inputs (finite approximation).
%    - Distinction: We adopt this execution-based principle for formal verification.
% ==================================================================

\section{Related Work}
\label{sec:related_work}

\paragraph{Formal Theorem Proving Benchmarks}
Recent benchmarks have driven progress in neural automated theorem proving.
MiniF2F~\cite{zheng2022minif2fcrosssystembenchmarkformal}, ProofNet~\cite{azerbayev2023proofnetautoformalizingformallyproving}, and PutnamBench~\cite{tsoukalas2024putnambenchevaluatingneuraltheoremprovers} evaluate models on high school Olympiad- to undergraduate-level problems, while Lean-Workbook~\cite{ying2025leanworkbooklargescalelean} and FormalMATH~\cite{yu2025formalmathbenchmarkingformalmathematical} provide large-scale training corpora. 
However, as summarized in \Cref{tab:contribution}, these benchmarks primarily verify logical validity via compilation, leaving semantic correctness either unaddressed or assessed through unreliable proxies such as BLEU, natural-language-inference-based back-translation, or prover-based equivalence checking, all of which require ground-truth references or external models and provide only indirect estimates of semantic correctness. 
Manual inspection~\cite{ospanov2025minif2fleanrevisitedreviewinglimitations} offers a reliable evaluation but lacks scalability.
Moreover, since these benchmarks consist of isolated, self-contained problems, they are far removed from real-world mathematical proof development, where theorems are deeply interconnected through dependency structures.

% \fix{
% \paragraph{Neural Theorem Provers and Formalizers}
% Beyond benchmarks, recent work has developed specialized systems for proof generation and autoformalization.
% DeepSeek-Prover \citep{ren2025deepseekproverv2} and Goedel-Prover \citep{lin2025goedelproverv2} leverage large-scale synthetic data and self-correction to scale formal reasoning, while Kimina-Prover \citep{wang2025kimina} applies reinforcement learning for proof generation.
% Autoformalization models such as Goedel-Formalizer \citep{lin2025goedelproverv2} and Kimina-Autoformalizer \citep{wang2025kimina} focus on translating natural language mathematics into formal statements.
% Building on these foundations, COPRA \citep{thakur2024copra} employs in-context learning to guide proof search in Lean and Coq, and LeanAgent \citep{kumarappan2024leanagent} introduces lifelong learning for continuously improving theorem proving across evolving repositories.
% \ours is complementary to these efforts: rather than improving proof generation or autoformalization, we provide an execution-based metric to evaluate the \emph{semantic correctness} of their outputs.
% }

\paragraph{Test-Based Evaluation in Code Generation}
In code generation, test-based evaluation is standard practice. 
Benchmarks such as APPS~\cite{DBLP:conf/nips/HendrycksBKMAGB21}, HumanEval~\cite{chen2021evaluatinglargelanguagemodels}, and MBPP~\cite{DBLP:journals/corr/abs-2108-07732} evaluate generated code by executing it against test cases rather than comparing it lexically to a reference solution. 
This paradigm directly measures functional correctness and is both automatic and reference-free. 
Despite the well-established connection between proofs and programs via the Curry-Howard correspondence~\cite{curry--howard_correspondence}, test-based evaluation has not been applied to formal theorem generation before our work.

\section{Testing-Based Semantic Evaluation}
\label{sec:metric_definition}

Evaluating semantic correctness, whether a generated theorem captures the intended logical meaning of an \gls{nl} theorem, is a central challenge in formal theorem generation.

% 없이도 설명되는 거 같아서 일단 빼두고 필요하면 넣기..
% \begin{figure}[t]
%     \centering
%     \includegraphics[width=\linewidth]{fig/categorization.pdf}
%     \caption{The relationship between Syntactic Validity, Proof Validity, and Semantic Correctness. 
%     Most existing metrics only check up to Proof Validity (compilation), leaving a gap for semantically incorrect proofs (e.g., tautologies) to pass.
%     }
%     \label{fig:validity_subset}
% \end{figure}

Given an \gls{nl} theorem $t_{\text{nl}}$, the task is to generate a formal theorem $t_{\text{fl}} = (s_{\text{fl}}, p_{\text{fl}})$ comprising a statement $s_{\text{fl}}$ and a proof $p_{\text{fl}}$, such that $t_{\text{fl}}$ is both logically valid and semantically faithful to $t_{\text{nl}}$.
% As described on \Cref{fig:validity_subset}, 
Compilation verifies \textbf{logical validity} ($s_{\text{fl}} \vdash p_{\text{fl}}$), but not \textbf{semantic correctness} ($s_{\text{fl}} \vdash p_{\text{fl}} \wedge t_{\text{fl}} \equiv \llbracket t_{\text{nl}} \rrbracket$). 
Proving strict logical equivalence ($t_{\text{fl}} \equiv t_{\text{GT}}$) is generally intractable for complex theories, and it is hard to find the ground-truth of the target theorem $t_{\text{GT}}$.
Existing proxies that rely on surface-level similarity or approximate entailment fail to capture the intended mathematical semantics, as stated in \Cref{sec:related_work}.

To address this, we adopt an observational approach: rather than proving equivalence directly, we verify whether $t_{\text{fl}}$ behaves equivalently to $t_{\text{GT}}$ in all observed contexts. 
% Drawing on execution-based software testing, we extend this principle to formal theorem generation via the Curry-Howard correspondence.

\subsection{Integration Testing for Formal Theorems}
\label{subsec:integration}
In software engineering, a function may be compiled yet fail to meet its specification. 
To catch such failures, integration testing validates that a component, when combined with its dependents, produces correct behavior \cite{ammann2017introduction}.
If a modified component causes dependent modules to fail, it signals a semantic violation.

\definecolor{predcolor}{RGB}{0,90,170}
\definecolor{targetcolor}{RGB}{180,30,30}        
\definecolor{succcolor}{RGB}{20,120,50}
\newcommand{\hlpred}[1]{\textcolor{predcolor}{\textbf{#1}}}
\newcommand{\hltgt}[1]{\textcolor{targetcolor}{\textbf{#1}}}
\newcommand{\hlsucc}[1]{\textcolor{succcolor}{\textbf{#1}}}

\usetikzlibrary{calc}
% \begin{figure}[t]
% \begin{lstlisting}[language=Lean, basicstyle=\ttfamily\scriptsize, escapeinside={(*@}{@*)}]
% -- Predecessors (*@$\mathcal{T}_{\text{pred}}$@*) 
% -- Predecessor Theorems: required to state (*@$t_{\text{GT}}$@*)
% def (*@\hlpred{InfiniteAdeleRing}@*) := (v : InfinitePlace K) -> v.completion
% ...

% -- Target (*@$t_{\text{GT}}$@*)
% -- Replaced by the generated (*@$t_{\text{fl}}$@*) during evaluation
% @[simp] theorem (*@\hltgt{algebraMap\_apply}@*) (x : K) :
%     algebraMap K (InfiniteAdeleRing K) x v = x := rfl

% -- Successors (*@$\mathcal{T}_{\text{succ}}$@*) 
% -- Successor Theorems test suite: depend on (*@$t_{\text{GT}}$@*)
% theorem (*@\hlsucc{mixedEmbedding\_eq\_algebraMap\_comp}@*) {x : K} :
%     mixedEmbedding K x = ringEquiv_mixedSpace K (algebraMap K (InfiniteAdeleRing K) x) := by
%   ext v <;> simp only [ringEquiv_mixedSpace_apply, algebraMap_apply, ...]
%   ...

% theorem (*@\hlsucc{ringEquiv\_mixedSpace\_apply}@*) (x : InfiniteAdeleRing K) :
%     ringEquiv_mixedSpace K x = ... := rfl
% \end{lstlisting}
% \caption{An example problem of \ours. The target $t_{\text{GT}}$ , \hltgt{\texttt{algebraMap\_apply}} has predecessors $\mathcal{T}_{\text{pred}}$, (e.g. definition of \hlpred{\texttt{InfiniteAdeleRing}}), and successors $\mathcal{T}_{\text{succ}}$ that invoke it  (e.g. \hlsucc{\texttt{mixedEmbedding\_eq\_algebraMap\_comp}}). 
% Substituting an incorrect $t_{\text{fl}}$ for $t_{\text{GT}}$ causes the successor theorems to fail compilation.}
% \label{fig:pred_succ_ex}
% \end{figure}

\begin{figure}[!h]
\centering

\begin{subfigure}[t]{\linewidth}
\begin{lstlisting}[language=Lean, basicstyle=\ttfamily\scriptsize, escapeinside={(*@}{@*)}]
-- Predecessors (*@$\mathcal{T}_{\text{pred}}$@*) 
-- Predecessor Theorems: required to state (*@$t_{\text{GT}}$@*)
def (*@\hlpred{InfiniteAdeleRing}@*) := (v : InfinitePlace K) -> v.completion
...

-- Target (*@$t_{\text{GT}}$@*)
-- Replaced by the generated (*@$t_{\text{fl}}$@*) during evaluation
@[simp] theorem (*@\hltgt{algebraMap\_apply}@*) (x : K) :
    algebraMap K ((*@\hlpred{InfiniteAdeleRing}@*) K) x v = x := rfl

-- Successors (*@$\mathcal{T}_{\text{succ}}$@*) 
-- Successor Theorems test suite: depend on (*@$t_{\text{GT}}$@*)
theorem (*@\hlsucc{mixedEmbedding\_eq\_algebraMap\_comp}@*) {x : K} :
    mixedEmbedding K x = ringEquiv_mixedSpace K (algebraMap K ((*@\hlpred{InfiniteAdeleRing}@*) K) x) := by
  ext v <;> simp only [ringEquiv_mixedSpace_apply, (*@\hltgt{algebraMap\_apply}@*), ...]
  ...

theorem (*@\hlsucc{ringEquiv\_mixedSpace\_apply}@*) (x : (*@\hlpred{InfiniteAdeleRing}@*) K) :
    ringEquiv_mixedSpace K x = ... := rfl
\end{lstlisting}
\caption{Lean source code}
\label{subfig:dependency_example_code}
\end{subfigure} 

\bigskip

\begin{subfigure}[t]{\linewidth}
\centering
\begin{tikzpicture}[
  every node/.style={font=\ttfamily\scriptsize, rounded corners, draw, very thick, inner sep=4pt, minimum height=1.6em},
  pred/.style={draw=predcolor, fill=predcolor!8, text=predcolor},
  tgt/.style={draw=targetcolor, fill=targetcolor!8, text=targetcolor},
  succ/.style={draw=succcolor, fill=succcolor!8, text=succcolor},
  arr/.style={->, thick, >=stealth, shorten >=2pt, shorten <=2pt},
  node distance=8mm and 4mm
]
\node[pred] (p1) {InfiniteAdeleRing};
\node[pred, right=of p1] (p2) {$\dots$};
\node[tgt, below=10mm of $(p1)!0.5!(p2)$] (t) {algebraMap$\dots$};
\node[succ, below left=10mm and -10mm of t] (s1) {mixedEmbedding$\dots$};
\node[succ, below right=10mm and -10mm of t] (s2) {ringEquiv$\dots$};
\node[succ, below=8mm of s1] (s1d) {$\dots$};
\node[succ, below=8mm of s2] (s2d) {$\dots$};

\draw[arr] (p1) -- (t);
\draw[arr] (p2) -- (t);
\draw[arr] (t) -- (s1);
\draw[arr] (t) -- (s2);
\draw[arr] (s1) -- (s1d);
\draw[arr] (s2) -- (s2d);

\node[draw=none, fill=none, left=2mm of p1, anchor=east, text=predcolor] {\textbf{$\mathcal{T}_{\text{pred}}$}};
\node[draw=none, fill=none, left=2mm of t, anchor=east, text=targetcolor] {\textbf{$t_{\text{GT}}$}};
\node[draw=none, fill=none, left=2mm of s1, anchor=east, text=succcolor] {\textbf{$\mathcal{T}_{\text{succ}}$}};
\end{tikzpicture}
\caption{Dependency graph}
\label{subfig:dependency_example_tree}
\end{subfigure}

\caption{An example problem of \ours. The target $t_{\text{GT}}$ (\hltgt{\texttt{algebraMap\_apply}}) has predecessors $\mathcal{T}_{\text{pred}}$, (e.g. definition of \hlpred{\texttt{InfiniteAdeleRing}}), and successors $\mathcal{T}_{\text{succ}}$ that invoke it  (e.g. \hlsucc{\texttt{mixedEmbedding\_eq\_algebraMap\_comp}}). 
% Substituting an incorrect $t_{\text{fl}}$ for $t_{\text{GT}}$ causes the successor theorems to fail compilation.
}
\label{fig:pred_succ_ex}
\end{figure}
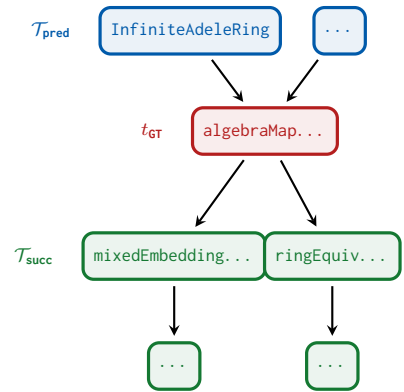
We adopt this principle for formal theorems by viewing each theorem as a 
node in a dependency graph \cite{prasad2024mathematical, cabral2026proofflow} rather than in isolation.
Let \hltgt{$t_{\text{GT}}$} be the \hltgt{ground-truth of the target theorem} corresponding to the \hltgt{generated candidate theorem $t_{\text{fl}}$}. 
We denote the theorems $t_{\text{GT}}$ depends on as 
\hlpred{predecessors $\mathcal{T}_{\text{pred}} = \{t_{\text{pred}}^{(1)}, \dots, t_{\text{pred}}^{(m)}\}$}, 
and the theorems that transitively depend on $t_{\text{GT}}$ as \hlsucc{successors 
$\mathcal{T}_{\text{succ}} = \{t_{\text{succ}}^{(1)}, \dots, t_{\text{succ}}^{(n)}\}$}, 
which serve as the test cases.
\Cref{subfig:dependency_example_code} shows a concrete Lean source view of these three types of theorem, where \hlpred{InfiniteAdeleRing} serves as a predecessor, \hltgt{algebraMap\_apply} is the target, and \hlsucc{mixedEmbedding\_eq\_algebraMap\_comp} is one of the successors that invokes the target. 
\Cref{subfig:dependency_example_tree} gives the corresponding dependency view, where predecessors flow into $t_{\text{GT}}$, and successors flow out of it. 
We consider $t_{\text{fl}}$ semantically correct 
if substituting it for $t_{\text{GT}}$\footnote{As a theorem with a different name will fail compilation easily, we condition the theorem name in the context.} leaves every theorem in 
$\mathcal{T}_{\text{succ}}$ compilable.

\subsection{Cut Elimination as Theorem Testing}
We ground this intuition formally in the Curry-Howard correspondence, 
% as summarized in \Cref{tab:theory\_correspondence}, 
which identifies proofs as programs and logical rules as computational operations. 
Specifically, we identify integration testing with the logical principle of \textit{Cut Elimination}.

The \textit{Cut Rule} in sequent calculus formalizes the composition of proofs. 
For instance, proving $X \to W$ by chaining intermediate theorems $Y$ and $Z$, given a shared set of background assumption $\Gamma$, corresponds to integration testing across multiple modules:
\begin{equation}
\label{eq:cut-rule}
    \frac{\Gamma \vdash X \to Y \quad \Gamma \vdash Y \to Z \quad \Gamma \vdash Z \to W}{\Gamma \vdash X \to W}
\end{equation}
where 
% $\Gamma$ corresponds to upstream theorems $\mathcal{T}_{up}$,
$X \to Y$ corresponds to $t_{\text{fl}}$, and the remaining chain represents $\mathcal{T}_{\text{succ}}$. 
In proof theory, the use of a lemma introduces a cut, and cut elimination removes the auxiliary lemma to produce a direct proof, which computationally corresponds to a function call. 
By the cut elimination theorem, if $t_{\text{fl}}$ is semantically correct, this chain of cuts can be eliminated, yielding a valid proof of $X \to W$. 
In other words, since $X \to W$ holds via the chain $X \to Y \to Z \to W$, if $t_{\text{fl}}$ breaks the compilation of $\mathcal{T}_{\text{succ}}$, then we treat $t_{\text{fl}}$ as semantically incorrect. Successful compilation across the chain provides evidence of semantic correctness, though not a logical guarantee.
Crucially, this process is not limited to immediate callers, so it can recursively verify $t_{\text{fl}}$ across the entire chain of dependent proofs.

\subsection{Testing Accuracy}
We apply the above principle to define our metric, \gls{ea}. 
Let $\mathcal{D}$ be the distribution of problems in the benchmark. 
For a problem $P = (t_{\text{nl}}, \mathcal{T}_{\text{pred}}, \mathcal{T}_{\text{succ}}) \in \mathcal{D}$, let $t_{\text{fl}}$ be the generated theorem. 
% Here, $\mathcal{T}_{\text{up}}$ corresponds to $\Gamma$ in \Cref{eq:cut-rule}, that is the upstream theorems on which both $t_{\text{fl}}$ and $\mathcal{T}_{\text{down}}$ utilize them.
Here, $\mathcal{T}_{\text{pred}}$ corresponds to $\Gamma$ in \Cref{eq:cut-rule}, namely the predecessor theorems used by both $t_{\text{fl}}$ and $\mathcal{T}_{\text{succ}}$.
Each successor theorem $t_{\text{succ}}^{(i)}$ imposes a semantic constraint on $t_{\text{fl}}$. 
As the number and depth of these constraints increase, their conjunction increasingly approximates a guarantee of semantic correctness.

\noindent
We define \gls{ea} as the expectation over problems:
\begin{equation}
    \texttt{TA} = \mathop{\mathbb{E}}_{P \sim \mathcal{D}} \left[ \bigwedge_{i=1}^{k} \texttt{compiles}(t_{\text{succ}}^{(i)} \mid t_{\text{fl}}) \right]
\end{equation}
where $\texttt{compiles}(t_{\text{succ}}^{(i)} \mid t_{\text{fl}}) = 1$ if $t_{\text{succ}}^{(i)}$ successfully compiles when $t_{\text{GT}}$ is replaced by $t_{\text{fl}}$, and $0$ otherwise.

% To ensure meaningful coverage, we restrict our benchmark to theorems with $|\mathcal{T}_{\text{down}}| \geq 2$, and our benchmark provides depth 7 and an average of 1.6k successor theorems per problem, requiring that each evaluation involves multiple semantic constraints (see \Cref{sec:benchmark}).
To ensure meaningful coverage, we restrict our benchmark to theorems with $|\mathcal{T}_{\text{succ}}| \geq 2$ (see \Cref{sec:benchmark}). 
The resulting problems have dependency depth up to 7 and an average of 1.6k successor theorems, so each evaluation involves multiple semantic constraints.

% 일단 빼두고 나중에 필요하면 넣기
% \begin{figure*}[t]
%     \centering
%     \includegraphics[width=0.8\linewidth]{fig/pipeline.pdf}
    
%     \caption{
%     \jy{update to small icons...}
%     The dataset extraction pipeline used to construct \ours. We begin by mining high-quality Lean 4 repositories to build a global dependency graph. We then select non-trivial targets with rich integration contexts and extract their upstream dependencies and successor test suites. Finally, we annotate these formal theorems with natural language descriptions using \glspl{llm}.
%     }
%     \label{fig:dataset_extraction}
% \end{figure*}

\section{Benchmark Construction}
\label{sec:benchmark}

% To support our proposed metric EA,
% we introduce \textbf{\ours} (\textbf{T}heorem \textbf{E}valuation by \textbf{E}xecution), a benchmark constructed by mining real-world Lean 4 projects for theorems 
% embedded in real dependency structures. 

% with rich successor dependencies. 
% we require theorems embedded in real dependency structures. 
% The pipeline proceeds as shown in \Cref{fig:dataset\_extraction}.

\subsection{Extraction Pipeline}
We employ a fully automated, dependency-aware extraction pipeline targeting public Lean 4 repositories.

\paragraph{Dependency Graph Construction}
We parse the Lean environment to build a global dependency graph $G=(V, E)$, where $V$ is the set of theorems and lemmas and $E$ represents usage relationships between them, with Lean-Dojo \cite{yang2023leandojo}.

\paragraph{Target Selection.}
From $G$, we select target theorems $t_{\text{GT}}$ that satisfy two criteria: 
(1) non-triviality: the depth, which is the distance from the target theorem to the successor theorems, is greater than 1 in $G$, excluding standalone theorems that existing benchmarks predominantly target, 
and (2) successor coverage: $|\mathcal{T}_{\text{succ}}| \geq 2$, ensuring that each target admits multiple semantic constraints for meaningful evaluation, reducing the possibility of false positives.

\paragraph{Context Extraction}
For each target $t_{\text{GT}}$, we extract the predecessor theorems $\mathcal{T}_{\text{pred}}$, consisting of the definitions and lemmas necessary to state and prove $t_{\text{GT}}$, and the successor theorems $\mathcal{T}_{\text{succ}}$, comprising all theorems that transitively depend on $t_{\text{GT}}$.

\subsection{Natural Language Annotation}
Since real-world repositories typically lack natural language descriptions, we generate $t_{\text{nl}}$ for each target using a strong \gls{llm} (Claude Sonnet 4.5). 
This allows us to construct a problem instance $P_i = (t_{\text{nl}},\, \mathcal{T}_{\text{pred}}(t_{\text{GT}}, t_{\text{succ}}^{(i)}),\, t_{\text{succ}}^{(i)})$ for each target $t_{\text{GT}}$ and each successor theorem $t_{\text{succ}}^{(i)} \in \mathcal{T}_{\text{succ}}(t_{\text{GT}})$.
To validate annotation quality, we manually verified 10 randomly sampled \gls{nl} annotations and found no errors. 
Full details are provided in Appendix~\ref{sec:appendix:nl-annotation}.

\subsection{Dataset Statistics}
We first examined whether existing benchmarks could support testing-based evaluation. 
However, few problems in these benchmarks satisfy our non-triviality and successor coverage criteria, yielding only 12 problems from evaluation benchmarks and 110 from training benchmarks (see \Cref{tab:dataset_stats}).

We therefore extract from 5 high-quality Lean 4 repositories listed on the Lean community page\footnote{\url{https://leanprover-community.github.io/papers.html}}. 
The resulting benchmark comprises 2,206 problems with an average of 41 successor theorems per target, providing a rigorous testbed for semantic correctness evaluation.
To further challenge model capabilities, we construct \ours Hard, a subset of 389 problems whose target declarations have a body of type Prop, requiring generation of both the proposition and its proof. 
% As shown in \Cref{tab:dataset_stats}, this subset is primarily drawn from \textit{math2001}, \textit{Directed-Topology}, and \textit{adele-ring}, and serves as our primary evaluation target throughout the experiments.
% ==================================================================
% Section 4: Experimental Setup
% ==================================================================
\section{Experiment Setup}
\label{sec:experimental_setup}

\subsection{Models}
We evaluate 18 models spanning general-purpose and domain-specialized systems, as summarized in \Cref{tab:models}. 
All models are evaluated in a zero-shot setting with temperature 0.6 and top\_p 0.95. 
Prompts are provided in \Cref{sec:appendix:prompts}.

\subsection{Metrics}
Our primary metric is \gls{ea}: the generated theorem $t_{\text{fl}}$ replaces the original theorem $t_{\text{GT}}$ in the repository, and we recompile the entire dependency chain as a test suite. 
The generated theorem passes the test if and only if all successor theorems compile successfully.
% The theorem is considered semantically correct if and only if all successor theorems compile successfully. 
\gls{ea} is the only metric that directly evaluates semantic correctness, and all main results are reported using it.

To contextualize our findings and demonstrate the limitations of existing evaluation methods, we additionally report two baseline metrics. 
Compilation accuracy verifies that the generated theorem compiles in isolation, corresponding to the standard pass criterion in existing ATP benchmarks.\footnote{For repository-sourced problems, we append \texttt{\#exit} after the generated theorem to prevent the Lean compiler from compiling successor theorems.} 
\textbf{BLEU} measures lexical similarity to the ground-truth via n-gram overlap. 
We analyze the gap between Compilation Accuracy and \gls{ea} to quantify how much compilation overestimates semantic correctness, and examine whether BLEU can serve as a reliable proxy for \gls{ea}, in \Cref{subsec:metric_comparison}.

\subsection{Evaluation Environment}
Our benchmark comprises two categories of problems with different verification requirements: \ours (including the \ours Hard subset).
\ours is drawn from real-world Lean repositories that rely on project-specific dependencies and varying Lean versions. 
For these problems, we execute \texttt{lake build} within each repository's native environment to strictly adhere to its original configuration and enable full dependency checking. 
For the supplementary existing benchmarks (e.g., FormalMath, CombiBench), which consist of single theorem problems without successor theorems, we use \texttt{kimina-lean-server}~\cite{santos2025kiminaleanservertechnical} to verify compilation within a standardized Lean 4.25.0 environment. 
Details are provided in \Cref{sec:appendix:dataset_details}.
\section{Results}
\label{sec:experiments}

We evaluate state-of-the-art \glspl{llm} on the \ours benchmark. 
We first present model performance and key results in \Cref{subsec:overview}. 
We then analyze the effect of few-shot prompting and iterative refinement with 
compiler feedback in \Cref{subsec:ablation:fewshot}, compare \gls{ea} 
against existing metrics in \Cref{subsec:metric_comparison}, and study the 
impact of successor theorem coverage in \Cref{subsec:downstream}.

\subsection{Overview of Results}
\label{subsec:overview}
\begin{table}[t]
    \centering
    \scalebox{0.7}{
        \renewcommand{\arraystretch}{1.1}
        \begin{tabular}{lcccc}
            \toprule
                                 & \multicolumn{2}{c}{\textbf{\ours}}                                     & \multicolumn{2}{c}{\textbf{\ours Hard}} \\
            \cmidrule(lr){2-3} \cmidrule(lr){4-5}
            
            \multicolumn{1}{c}{\textbf{Model}}           & \multicolumn{1}{c}{\textbf{Compile}} & \multicolumn{1}{c}{\textbf{TA}} & \multicolumn{1}{c}{\textbf{Compile}} & \multicolumn{1}{c}{\textbf{TA}} \\
            \midrule
            \multicolumn{5}{c}{\textit{\textbf{Closed-Source LLMs}}} \\
            \midrule
            Claude-3.7-Sonnet        & 75.1 & 36.8 & 34.3 & 2.6 \\
            Claude-4-Sonnet          & 81.1 & 36.0 & 47.5 & 4.5 \\
            Claude-Sonnet-4.5        & 80.3 & 38.9 & 46.0 & 4.5 \\
            GPT-4o-mini              & 59.1 & 36.7 & 12.2 & 1.5 \\
            GPT-5                    & 85.7 & 37.7 & 68.3 & 3.4 \\
            GPT-5-mini               & 86.0 & 37.9 & 66.6 & 4.1 \\
            GPT-5-nano               & 88.7 & 36.6 & 75.6 & 1.5 \\
            \midrule
            \multicolumn{5}{c}{\textit{\textbf{Open-Source LLMs}}} \\
            \midrule
            DeepSeek-R1              & 70.5 & 36.3 & 25.1 & 4.1 \\
            GPT-OSS-120B             & 68.2 & 32.3 & 40.0 & 4.5 \\
            Llama-3.1-405B           & 63.4 & 33.2 & 27.8 & 2.6 \\
            Llama-3.1-70B            & 65.4 & 37.0 & 20.1 & 2.4 \\
            Llama-3.1-8B             & 36.9 & 24.8 & 6.4 & 1.3 \\
            \midrule
            \multicolumn{5}{c}{\textit{\textbf{Math Specialized Models}}} \\
            \midrule
            DeepSeek-Prover-v2-7B    & 62.2 & 30.0 & 35.5 & 3.2 \\
            Goedel-Formalizer-v2-8B  & 38.7 & 23.6 & 22.1 & 2.4 \\
            Goedel-Prover-v2-32B     & 74.6 & 31.0 & 54.6 & 3.2 \\
            Goedel-Prover-v2-8B      & 46.1 & 24.5 & 37.9 & 2.6 \\
            Kimina-Autoformalizer-7B & 21.9 & 20.0 & 4.3 & 1.5 \\
            Kimina-Prover-Distill-8B & 28.0 & 23.4 & 5.1 & 1.7 \\
            \bottomrule
        \end{tabular}
    }
    \caption{
    Compilation accuracy and TA (\%) on \ours and \ours Hard.
    }
    \label{tab:main_table_ours_only}
\end{table}

\Cref{tab:main_table_ours_only} presents compilation accuracy and \gls{ea} across \ours and \ours Hard subsets, under the default setting (providing both \gls{nl} proof and successor theorem).
The best model (Claude-Sonnet-4.5) achieves 80.3\% compilation accuracy on the Full set while achieving only 38.9\% \gls{ea}.
Scaling model size yields modest improvements within model families (e.g., Llama-3.1 improves from 24.8\% at 8B to 33.2\% at 405B), but overall \gls{ea} remains below 40\% (Full) and 6\% (Hard), indicating that the semantic gap is far from closed.
We additionally evaluate on existing benchmarks, where the gap between compilation and \gls{ea} is substantially smaller due to the limited number of successor theorems.
Full results are provided in \Cref{sec:appendix:detailed_results}.

\begin{table}[t]
    \centering
    \scalebox{0.7}{
        \renewcommand{\arraystretch}{1.1}
        \begin{tabular}{l *{4}{>{\centering\arraybackslash}p{1.1cm}}} 
            \toprule
             & \multicolumn{2}{c}{\textbf{NL \cmark}} & \multicolumn{2}{c}{\textbf{NL \xmark}} \\
            \cmidrule(lr){2-3} \cmidrule(lr){4-5}
            \multicolumn{1}{c}{\textbf{Model}} & \textbf{ST \xmark} & \textbf{ST \cmark} & \textbf{ST \xmark} & \textbf{ST \cmark} \\
            \midrule
            \multicolumn{5}{c}{\textit{\textbf{Closed-Source LLMs}}} \\
            \midrule
            Claude-3.7-Sonnet        & 29.4 & 36.8 & 29.8 & 32.4 \\
            Claude-4-Sonnet          & 27.7 & 36.0 & 30.0 & 32.3 \\
            Claude-Sonnet-4.5        & 33.0 & 38.9 & 34.0 & 32.9 \\
            GPT-4o-mini              & 33.8 & 36.7 & 26.2 & 26.7 \\
            GPT-5                    & 36.9 & 37.7 & 29.1 & 28.7 \\
            GPT-5-mini               & 35.2 & 37.9 & 29.6 & 30.9 \\
            GPT-5-nano               & 33.4 & 36.6 & 30.8 & 31.1 \\
            \midrule
            \multicolumn{5}{c}{\textit{\textbf{Open-Source LLMs}}} \\
            \midrule
            DeepSeek-R1              & 31.0 & 36.3 & 28.9 & 29.1 \\
            GPT-OSS-120B             & 26.7 & 32.3 & 28.1 & 27.4 \\
            Llama-3.1-405B           & 30.8 & 33.2 & 29.7 & 30.8 \\
            Llama-3.1-70B            & 28.9 & 37.0 & 28.5 & 29.1 \\
            Llama-3.1-8B             & 22.6 & 24.8 & 21.6 & 21.8 \\
            \midrule
            \multicolumn{5}{c}{\textit{\textbf{Math Specialized Models}}} \\
            \midrule
            DeepSeek-Prover-v2-7B    & 32.0 & 30.0 & 28.2 & 26.1 \\
            Goedel-Formalizer-v2-8B  & 23.3 & 23.6 & 24.3 & 23.6 \\
            Goedel-Prover-v2-32B     & 30.6 & 31.0 & 29.4 & 31.3 \\
            Goedel-Prover-v2-8B      & 25.4 & 24.5 & 24.7 & 24.2 \\
            Kimina-Autoformalizer-7B & 20.2 & 20.0 & 20.0 & 19.9 \\
            Kimina-Prover-Distill-8B & 21.3 & 23.4 & 20.7 & 22.2 \\
            \bottomrule
        \end{tabular}%
    }
    \caption{
    Impact of input context on \gls{ea} (\%) on \ours full dataset. 
    We compare two axes: whether an \gls{nl} proof is provided (\cmark) and whether a successor theorem (ST) is provided (\cmark) or withheld (\xmark).
    }
    \label{tab:ablation_research_full}
\end{table}
We further investigate the impact of input context on \gls{ea}. \Cref{tab:ablation_research_full} presents the full ablation on the \ours dataset. 
We find that both \gls{nl} proofs and successor theorems must be provided together to yield consistent improvements, that is, neither alone offers meaningful benefit. 
For instance, Claude-4.5-Sonnet improves by 4.9\% when given both \gls{nl} proofs and successor theorems, yet providing successor theorems alone without \gls{nl} proofs actually decreases \gls{ea}. 
This suggests that the successor theorem helps the model understand the required semantic constraints, while \gls{nl} proofs provide the necessary reasoning context to satisfy them.
The ablation results on \ours Hard subset are provided in \Cref{sec:appendix:ablation}.

\subsection{Extended Experiments: Few-shot and Iterative Refinement}
\label{subsec:ablation:fewshot}

To test whether other prompting strategies can improve semantic correctness, we evaluate 2-shot prompting and iterative refinement on the \ours Hard subset.

\paragraph{Few-shot prompting}
We test whether the low performance on \ours Hard comes from the models struggling to understand the task format, or from a gap in their ability to solve it. 
We evaluate the four top-performing models with 2-shot prompting. 
We sample two examples from the Herald~\cite{herald} training set that Claude Sonnet 4.5 solves, and prepend them as in-context examples.
\Cref{tab:fewshot} compares \gls{ea} between the zero-shot and 2-shot settings. 
We see no consistent gain across the four models, and in some cases, the performance drops. 
This shows that the difficulty of \ours does not come from understanding what the task asks for. 
The harder part is producing a formalization that matches the natural-language annotation in meaning, which a few examples cannot teach.

\begin{table}[t]
% \centering
% \small
\scalebox{0.8}{
\renewcommand{\arraystretch}{1.1}
\begin{tabular}{lcc}
\toprule
\textbf{Model} & \textbf{Zero-shot TA (\%)} & \textbf{2-shot TA (\%)} \\
\midrule
Claude-4-Sonnet & 4.5 & 4.3 \\
Claude-3.7-Sonnet & 2.6 & 2.8 \\
GPT-5 & 3.4 & 2.6 \\
GPT-5-mini & 4.1 & 2.8 \\
\bottomrule
\end{tabular}
}
\caption{Zero-shot vs.\ 2-shot \gls{ea} on \ours Hard.}
\label{tab:fewshot}
\end{table}

\paragraph{Iterative refinement}
We also evaluate Hilbert~\citep{varambally2025hilbert}, a method that takes feedback from the Lean compiler and uses it to refine the generated theorems over multiple rounds. Even with this feedback loop, Hilbert reaches 5.0\% \gls{ea} on \ours Hard, compared to 3.2\% from the zero-shot baseline (DeepSeek-Prover-v2-7B). 
The gain is small and does not reduce the semantic gap.

% \fix{
% We additionally evaluated few-shot prompting and iterative refinement with compiler feedback on \ours Hard and found no meaningful improvement as described in \Cref{sec:appendix:fewshot}.
% }

\subsection{Comparison with Existing Metrics}
\label{subsec:metric_comparison}

\paragraph{Compilation rate}
\begin{figure}[!t]
    \centering
    \includegraphics[width=0.9\linewidth]{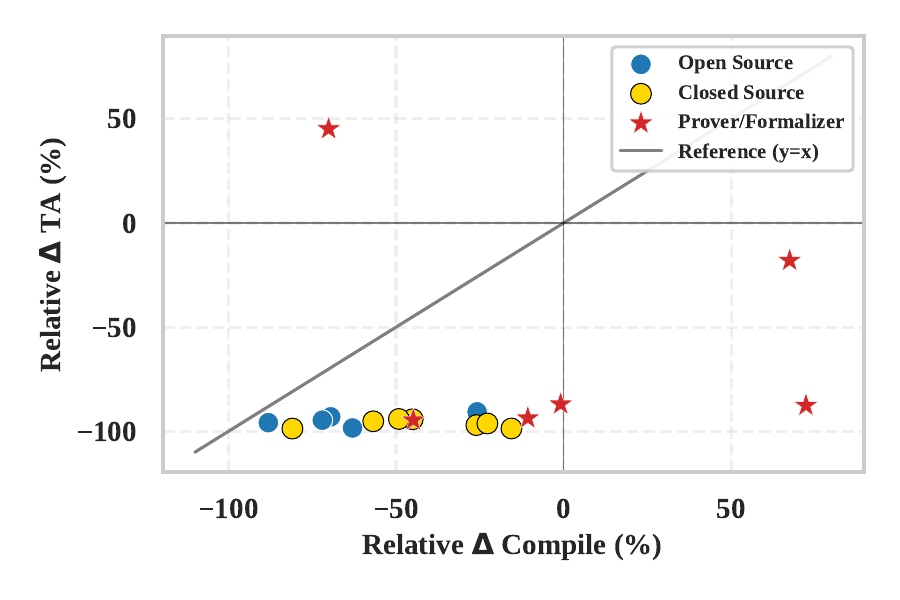}
    \caption{
    Relative performance drop from existing benchmarks to \ours Hard, in compilation accuracy (x-axis) vs.\ \gls{ea} (y-axis). 
    Prover/Formalizer models (\textcolor{red}{red} stars) in the lower-right quadrant, where compilation gains do not translate to \gls{ea} improvements.
    % indicating syntactic fluency without semantic understanding.
    }
    \label{fig:compile_vs_ea}
\end{figure}

As shown in \Cref{tab:main_table_ours_only}, there is a significant divergence between compilation success and semantic correctness across all models. 
The precision of compilation as a predictor of semantic correctness is only 6.89\%.
For instance, GPT-5 achieves \textbf{68.3\%} compilation accuracy but only \textbf{3.4\%} \gls{ea}. 
This disparity persists regardless of model scale or specialization, confirming that compilation metrics fail to capture semantic correctness. 
Notably, as shown in \Cref{fig:compile_vs_ea}, specialized proving models report compilation gains over general-purpose models, but these gains do not translate to \gls{ea} improvements, suggesting that domain-specific fine-tuning improves syntactic fluency without improving semantic correctness.

\paragraph{Lexical metric}
\begin{figure}[!t]
    \centering
    \includegraphics[width=\linewidth]{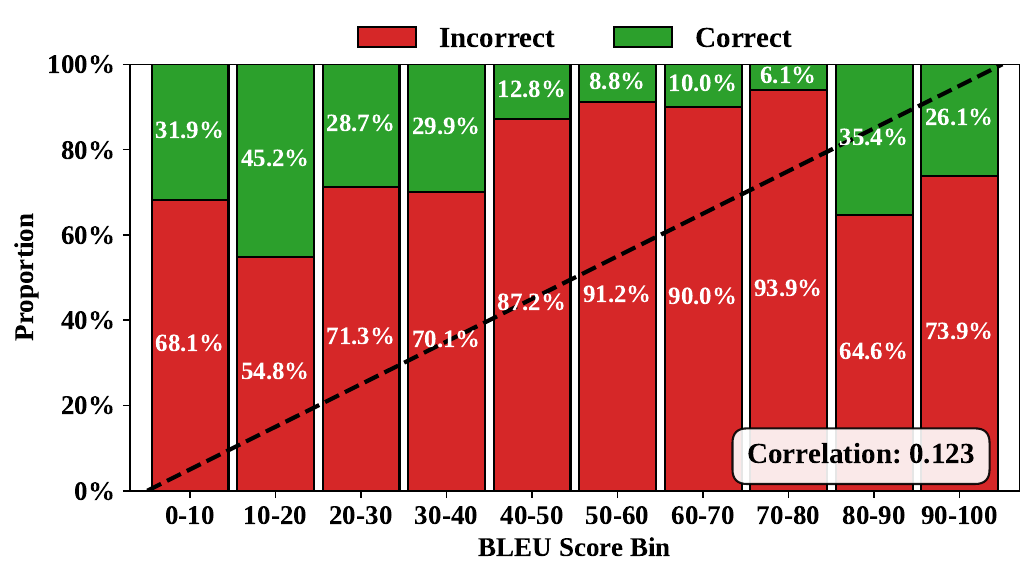}
    \caption{
    Proportion of testing-verified (green) and testing-failed (red) samples across BLEU score bins, aggregated over all models. High BLEU scores do not guarantee semantic correctness.
    }
    \label{fig:bleu_bar}
\end{figure}

We find that lexical similarity is an unreliable proxy for semantic correctness. 
As shown in \Cref{fig:bleu_bar}, even in the highest BLEU score bins (90-100), over 70\% of samples are semantically incorrect. 
Conversely, the lowest bins (0-10) still contain approximately 30\% verified correct samples. 
This demonstrates that high BLEU scores do not guarantee semantic correctness, nor do low scores, confirming that lexical metrics cannot be utilized for semantic correctness.

\subsection{Analysis of Successor Theorem Coverage}
\label{subsec:downstream}

\begin{figure}[t]
\centering
\begin{subfigure}[b]{0.49\linewidth}
\centering
\includegraphics[width=\linewidth]{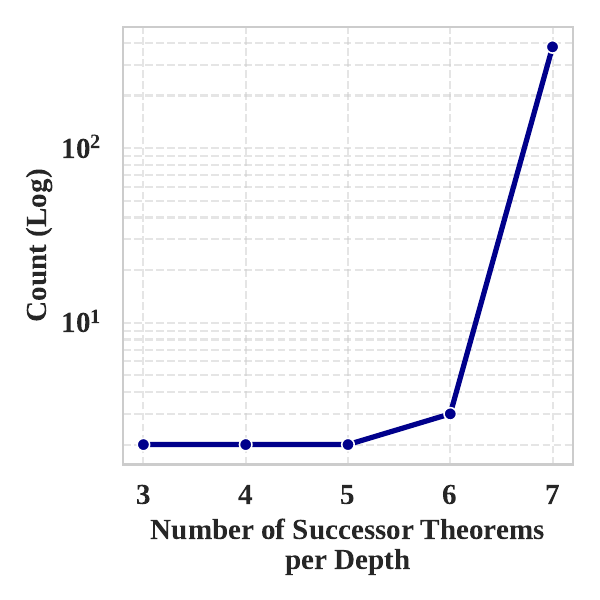}
\caption{Depth distribution}
\label{fig:depth_count}
\end{subfigure}
\begin{subfigure}[b]{0.49\linewidth}
\centering
\includegraphics[width=\linewidth]{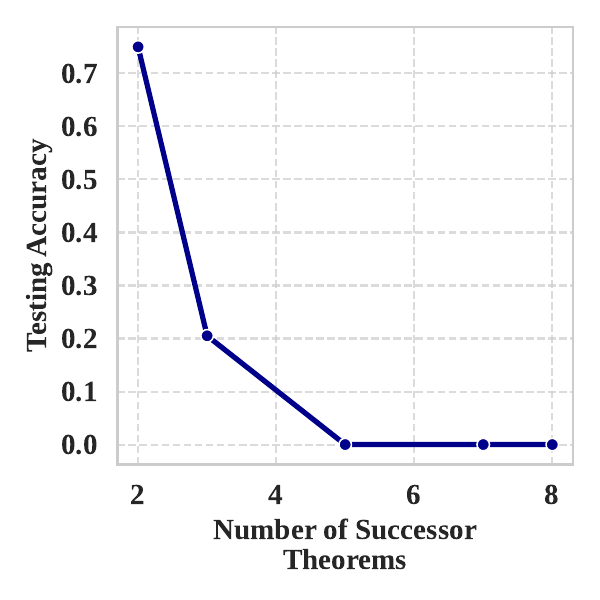}
\caption{\gls{ea} by number of STs}
\label{fig:width_analysis}
\end{subfigure}
\caption{
% (a) Distribution of successor theorem tree depth in \ours Hard. 
% The majority of problems (380) reach depth 7, with an average of 1,600 successor theorems (STs) at that depth. 
% (b) \gls{ea} in \ours Hard set for Claude-4-Sonnet decreases sharply as the number of successor theorems increases, demonstrating that broader dependency coverage provides stricter semantic evaluation.
(a) Distribution of successor theorem (ST) depth in \ours Hard. Most problems (380) reach depth 7, with 1,600 STs on average at that depth.
(b) \gls{ea} drops sharply as STs grow, indicating that broader coverage yields stricter evaluation.
}
\label{fig:dt_count}
\end{figure}

We further examine how successor theorem coverage affects evaluation strictness. 
As shown in \Cref{fig:depth_count}, the majority of problems in our benchmark have a dependency depth of 7, ensuring deep semantic verification by default. 
\Cref{fig:width_analysis} shows that \gls{ea} decreases sharply as the number of successor theorems increases: with only 2 successor theorems, Claude-4-Sonnet achieves approximately 70\% \gls{ea}, dropping to near 0\% when 5 or more must be satisfied. 
This confirms that each additional successor theorem imposes a meaningful semantic constraint, and that our benchmark's rich dependency structures provide a rigorous evaluation setting.
\section{Conclusion}
\label{sec:conclusion}
We introduced \gls{ea}, the first testing-based metric for evaluating semantic correctness in automated theorem proving, and \ours, a large-scale Lean benchmark of 2,206 theorems with rich dependency structures. 
Our experiments across 18 models reveal that compilation-based evaluation fundamentally overestimates model capability, with up to 2$\times$ gap between compilation and \gls{ea} on the full set and up to 50$\times$ on the hard set. 
We further showed that lexical metrics such as BLEU cannot reliably distinguish semantically correct theorems from incorrect ones, and that providing successor theorems as generation context consistently improves semantic correctness. 
These findings demonstrate that testing-based evaluation is essential for trustworthy assessment of formal theorem generation.

\section{Limitations}
\label{sec:limitations}

The reliability of \gls{ea} increases with the number and diversity of successor theorems, as each imposes an additional semantic constraint.
However, for problems with few successor theorems, coverage may be insufficient to catch all semantic errors, and even with many dependents, full completeness cannot be guaranteed, analogous to the inherent limitation of software testing.
\gls{ea} is inherently inapplicable to standalone theorems without dependents, such as the most recently added theorems at the frontier of a repository that no other proof yet builds upon.
Both our metric and benchmark are implemented exclusively for Lean~4.
Moreover, \ours is constructed from 5 Lean~4 repositories covering primarily research-level mathematics.
The \gls{nl} specifications $t_{\text{nl}}$ are generated by a strong LLM rather than written by the original authors.
While this enables scalable annotation, it may introduce noise that affects the autoformalization setting.

\section*{Acknowledgments}

This work was supported by 
% autocoding
Electronics and Telecommunications Research Institute (ETRI) grant funded by ICT R\&D program of MSIT/IITP (2022-0-00995, Automated reliable source code generation from natural language descriptions),
% BRL
the National Research Foundation of Korea(NRF) grant funded by the Korea government(MSIT) (No. RS-2024-00414981), 
% 상식추론
% Institute of Information \& communications Technology Planning \& Evaluation (IITP) grant funded by the Korea government (MSIT) (No. 2022-0-00077/RS-2022-II220077, AI Technology Development for Commonsense Extraction, Reasoning, and Inference from Heterogeneous Data)
and 
% GSP
Institute of Information \& communications Technology Planning \& Evaluation (IITP) grant funded by the Korea government(MSIT) [NO.RS-2021-II211343, Artificial Intelligence Graduate School Program (Seoul National University)].

\bibliography{anthology,custom}

\clearpage
\appendix
% \section{Appendix}
\crefalias{section}{appendix}
\label{sec:appendix}

\begin{figure}[t]
    \centering
    \includegraphics[width=\linewidth]{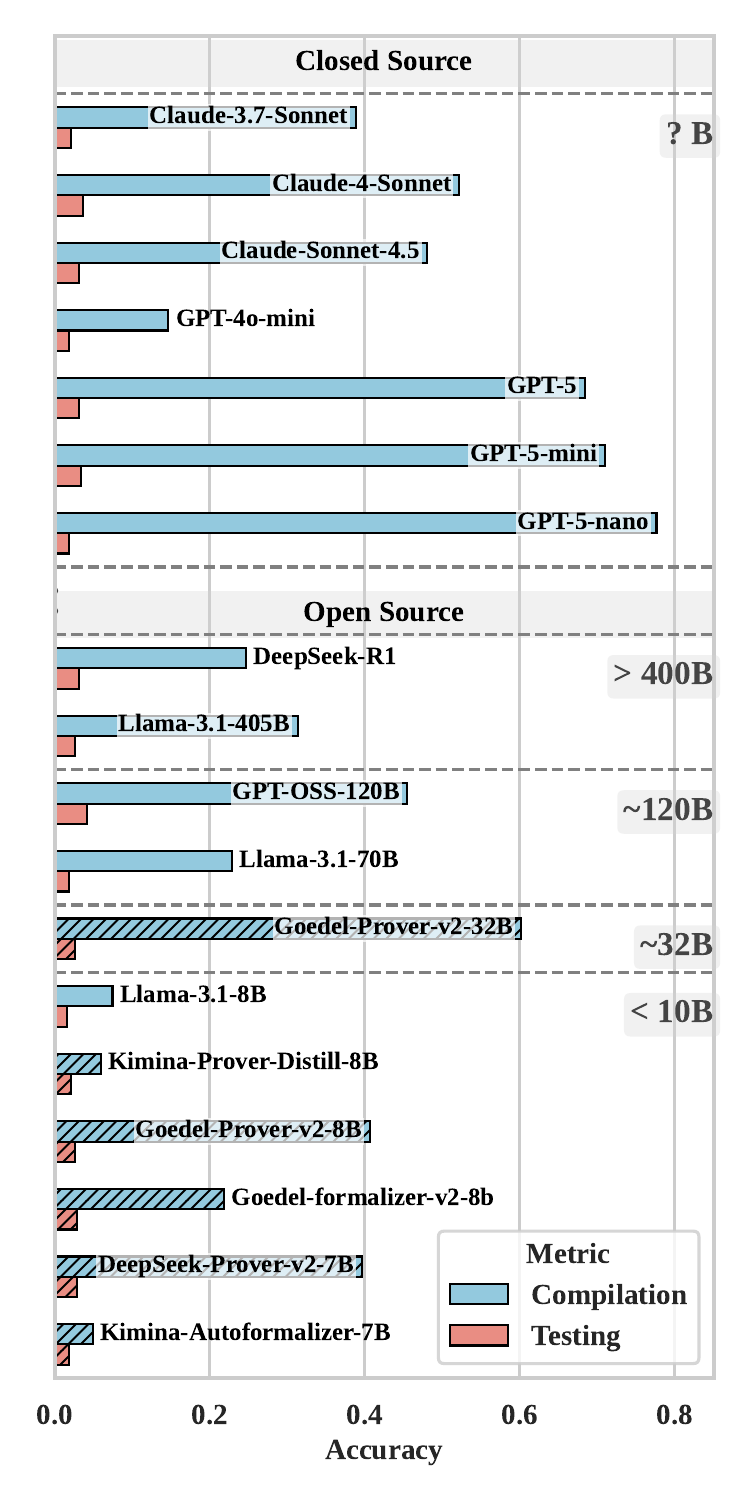}
    \caption{
    Compilation accuracy and testing accuracy on \ours Hard. 
    Models are grouped by source availability and parameter scale. 
    Dashed bars indicate \glspl{llm} fine-tuned on formal mathematical reasoning tasks.
    }
    \label{fig:model_size_scaling}
\end{figure}

\section{Model Details}
\label{sec:appendix:model_details}
\begin{table}[h]
  \centering
  \scriptsize
  \tabcolsep=4pt
  \renewcommand{\arraystretch}{1.1}
  \begin{tabular}{l l}
    \toprule
    \textbf{Family} & \textbf{Models} \\
    \midrule
    \multirow{3}{*}{Claude~\cite{claude37sonnet,claude3model,claude45sonnet}}
      & Claude-3.7-Sonnet \\
      & Claude-4-Sonnet \\
      & Claude-Sonnet-4.5 \\
    \midrule
    \multirow{4}{*}{GPT~\cite{gpt4omini,gpt5}}
      & GPT-4o-mini \\
      & GPT-5 \\
      & GPT-5-Mini \\
      & GPT-5-Nano \\
    \midrule
    GPT-OSS~\cite{gptoss120b} & GPT-OSS-120B \\
    \midrule
    \multirow{3}{*}{Llama~\cite{grattafiori2024llama}}
      & Llama-3.1-8B \\
      & Llama-3.1-70B \\
      & Llama-3.1-405B \\
    \midrule
    DeepSeek~\cite{deepseekr1} & DeepSeek-R1 \\
    \midrule
    DeepSeek-Prover~\cite{deepseekproverv2} & DeepSeek-Prover-v2-7B \\
    \midrule
    \multirow{3}{*}{Goedel-Prover~\cite{goedelproverv2}}
      & Goedel-Prover-v2-8B \\
      & Goedel-Prover-v2-32B \\
      & Goedel-formalizer-v2-8B \\
    \midrule
    \multirow{2}{*}{Kimina~\cite{kimina}}
      & Kimina-Autoformalizer-7B \\
      & Kimina-Prover-Distill-8B \\
    \bottomrule
  \end{tabular}
  
  \caption{
  Models evaluated in our experiments.
  }
  \label{tab:models}
\end{table}
We evaluate 18 models across 8 model families, as listed in \Cref{tab:models}.
\textbf{Closed-source models} are accessed via Snowflake AI Cortex API.
Parameter counts for these models are not publicly disclosed, except for GPT-OSS-120B.
\textbf{Open-source models} are served locally using vLLM.
The Llama-3.1 family (8B, 70B, 405B) and DeepSeek-R1 are general-purpose models, while DeepSeek-Prover-v2-7B, Goedel-Prover-v2 (8B, 32B), Goedel-formalizer-v2-8B, Kimina-Autoformalizer-7B, and Kimina-Prover-Distill-8B are fine-tuned on formal mathematical corpora for theorem proving or autoformalization tasks.
All models are evaluated in a zero-shot setting with temperature 0.6 and top\_p 0.95.
For each problem, we generate a single completion and evaluate it against the successor theorems as a test suite.
Closed-source model evaluations were conducted between January and February 2026.

\section{Dataset Details}
\label{sec:appendix:dataset_details}
\begin{table*}[t]
    \centering
    \scalebox{0.85}{
        \begin{tabular}{l c c c c}
            \toprule
            \multicolumn{5}{c}{\textbf{Overall Statistics}}                                                                                                    \\
            \midrule
            \multirow{2}{*}{\textbf{Benchmark}} & \multirow{2}{*}{\textbf{\# Task}} & \textbf{Test (Avg.)} & \textbf{Prompt (Avg.)} & \textbf{Solution (Avg.)} \\
            \cmidrule(lr){3-3} \cmidrule(lr){4-4} \cmidrule(lr){5-5}
                                                &                                   & \#                   & Char.                  & Char.                    \\
            \midrule
            \multicolumn{5}{l}{\textit{Existing Benchmarks}}                                                                                                     \\
            Kimina (train)                      & 5                                 & 2.00                 & 1208.00                & 48.00                    \\
            Numina (train)                      & 105                               & 6.53                 & 1185.65                & 114.01                   \\
            combibench (eval)                   & 6                                 & 2.67                 & 1229.83                & 70.33                    \\
            formalmath (eval)                   & 6                                 & 13.00                & 217.00                 & 66.17                    \\
            \cmidrule(lr){1-5}
            \textit{Total / Avg.}               & \textbf{122}                      & \textbf{6.48}        & \textbf{1141.10}       & \textbf{106.80}          \\
            \midrule
            \multicolumn{5}{l}{\textit{\ours}}                                                                                                          \\
            Directed-Topology                   & 438                               & 19.68                & 5489.52                & 172.28                   \\
            Untangle                            & 133                               & 16.71                & 2413.51                & 727.90                   \\
            WhitneyGraustein                    & 4                                 & 2.00                 & 177.75                 & 22.00                 
            \\
            % Duper-ITP                    & 4                                 & 2.00                 & 177.75                 & 22.00                 
            % \\
            adele-ring                          & 67                                & 4.52                 & 3548.19                & 270.45                   \\
            math2001                            & 1564                              & 50.12                & 1056.79                & 54.53                    \\
            \cmidrule(lr){1-5}
            \textit{Total / Avg.}               & \textbf{2206}                     & \textbf{40.59}       & \textbf{2092.78}       & \textbf{125.00}          \\
            \midrule
            \multicolumn{5}{l}{\textit{\ours (Hard)}}                                                                                                     \\
            Directed-Topology                   & 6                                 & 3.00                 & 5360.17                & 42.00                    \\
            adele-ring                          & 17                                & 2.53                 & 3239.24                & 337.76                   \\
            math2001                            & 366                               & 6.97                 & 1476.30                & 45.97                    \\
            \cmidrule(lr){1-5}
            \textit{Total / Avg.}               & \textbf{389}                      & \textbf{6.71}        & \textbf{1613.25}       & \textbf{58.66}           \\
            \bottomrule
        \end{tabular}
    }
    \caption{
    Dataset statistics for \ours benchmark, including the number of tasks, average successor theorems (Test), and average prompt/solution lengths in characters.
    Untangle and WhitneyGraustein are absent in Hard set as they have no problem matches Hard set criteria.
    }
    \label{tab:dataset_stats}
\end{table*}
\begin{table*}[t]
    \centering
    \scalebox{0.8}{
        \begin{tabular}{l l c c}
            \toprule
            \textbf{Repository} & \textbf{Github Link}                                    & \textbf{Lean Version} & \textbf{Difficulty} \\
            \midrule
            adele-ring          & \url{smmercuri/adele-ring_locally-compact/tree/journal} & v4.10.0-rc2           & Research            \\
            Directed-Topology   & \url{Dominique-Lawson/Directed-Topology-Lean-4}         & v4.6.0-rc1            & Research            \\
            % Duper\_ITP          & \url{JOSHCLUNE/Duper_ITP_Paper_Artifact}                & v4.6.0-rc1            & Research            \\
            Untangle            & \url{dignissimus/Untangle}                              & v4.6.1                & Research            \\
            math2001            & \url{hrmacbeth/math2001}                                & v4.3.0                & Undergraduate       \\
            WhitneyGraustein    & \url{MetalCreator666/WhitneyGraustein}                  & v4.10.0               & Research            \\
            \bottomrule
        \end{tabular}
    }
    \caption{Details of the Source Repositories used in \ours}
    \label{tab:repository_detail}
\end{table*}

We constructed our \ours benchmark by mining high-quality, real-world Lean 4 repositories.
\Cref{tab:repository_detail} lists the source projects included in our dataset, covering a diverse range of mathematical domains such as algebra, topology, and number theory.
This diversity is crucial for evaluating the generalization capabilities of autoformalization models beyond standard educational problems.
For evaluation, we employ a rigorous environment setup using \code{lake build} within each repository to ensure that all generated theorems are checked against the correct dependencies and definitions.

\section{NL Annotation Validation}
\label{sec:appendix:nl-annotation}
To validate the quality of LLM-generated \gls{nl} annotations, we randomly sampled 10 annotations from \ours and manually verified them against the corresponding formal theorems.
Each annotation was checked for (1) mathematical accuracy of the described statement, (2) completeness of hypotheses and conclusions, and (3) absence of hallucinated or missing conditions.
All 10 annotations were judged correct, as summarized in \Cref{tab:nl_validation}.

\begin{table*}[h]
\centering
\small
\begin{tabular}{llc}
\toprule
\textbf{Theorem Name} & \textbf{Repository} & \textbf{Verification} \\
\midrule
\texttt{isUniformizer\_ne\_zero} & adele-ring & Correct \\
\texttt{extensionEmbedding\_of\_comp\_coe} & adele-ring & Correct \\
% \texttt{σCCWPoints.gp} & EmptyHexagon & Correct \\
% \texttt{restoreAtDep\_zero} & Duper & Correct \\
% \texttt{HList.reverseAux\_eq\_append} & Duper & Correct \\
\texttt{isometry\_extensionEmbedding\_of\_comp} & adele-ring & Correct \\
\texttt{algebraMap\_injective} & adele-ring & Correct \\
% \texttt{gp} (CanonicalPoints) & EmptyHexagon & Correct \\
\texttt{eq\_pow\_uniformizer\_mul\_unit} & adele-ring & Correct \\
\texttt{not\_isUnit\_iff\_valuation\_lt\_one} & adele-ring & Correct \\
\texttt{bijective\_extensionEmbedding\_of\_isReal} & adele-ring & Correct \\
\texttt{outputEnd} & Untangle & Correct \\
\texttt{Int.ModEq} & math2001 & Correct \\ 
% \texttt{saturate} & Duper & Correct \\
% \texttt{Array.finRange} & EmptyHexagon & Correct \\
\texttt{parseExpression} & Untangle & Correct \\
\bottomrule
\end{tabular}
\caption{Manual verification of 10 randomly sampled NL annotations. All annotations were judged to be mathematically accurate.}
\label{tab:nl_validation}
\end{table*}

\noindent
For a qualitative example, consider the following annotation:

\begin{tcolorbox}[title=Formal Theorem, colback=gray!5, colframe=gray!50, fonttitle=\bfseries\small]
\begin{lstlisting}[language=lean]
theorem EmptyShapeIn.rfl {S : Set Point} :
    EmptyShapeIn S S := by
  intro _ h
  simp at h
\end{lstlisting}
\end{tcolorbox}

\begin{tcolorbox}[title=Generated NL Annotation, colback=blue!3, colframe=blue!40, fonttitle=\bfseries\small]
For any set of points $S$, the set $S$ carves out an empty shape in itself. That is, the convex hull of $S$ contains no points from $S$ other than those already in $S$.
\end{tcolorbox}

\noindent
This NL description contains some redundancy (no points from $S$ other than those already in $S$), but it correctly captures the semantics of the formal statement. 
Such minor stylistic issues do not affect the autoformalization task, as the logical content is faithfully conveyed.

\section{Evaluation Environment Details}
\label{sec:appendix:eval_environment}
All experiments are conducted on a server equipped with 2 NVIDIA A6000 or 1 NVIDIA 6000Pro GPUs, having 49 GB and 98 GB VRAM, respectively.
Open-source models are served using vLLM with bfloat16 precision.
For models exceeding single-GPU memory (e.g., Llama-3.1-405B, DeepSeek-R1), we use tensor parallelism across 2 GPUs for NVIDIA A6000.

For \ours problems, we execute \texttt{lake build} within each repository's native environment. We set a timeout of 600 seconds per problem for both theorem generation and compilation verification.
For Standalone benchmarks like supplementary existing benchmarks, we use \texttt{kimina-lean-server} within a standardized Lean 4.25.0 environment with a timeout of 120 seconds.

\section{Computation Cost}
\label{sec:appendix:computation_cost}
\Cref{tab:computation_cost} summarizes the approximate computation cost of our experiments.

\begin{table*}[h]
\centering
\small
\begin{tabular}{lrr}
\toprule
Category & Time & Cost (USD) \\
\midrule
Closed-source API calls & total 12 hrs & approx.\ 2500 \\
Open-source inference & up to 48 GPU-hrs & -- \\
Compilation verification & 96 hrs & -- \\
NL annotation & 1 hr & approx.\ 100 \\
\midrule
Total & approx.\ 157 hrs & 2600 \\
\bottomrule
\end{tabular}
\caption{Approximate computation cost of experiments.}
\label{tab:computation_cost}
\end{table*}

The dominant cost is closed-source API calls, as each of the 389 Hard problems (and 2,206 full problems) requires a separate generation call per model across 4 context settings.
Compilation verification is comparatively inexpensive, as \texttt{lake build} runs on a CPU.

\section{Ablation Study: Input Context}
\label{sec:appendix:ablation}

\begin{table}[t]
    \centering
    \scalebox{0.7}{
        \renewcommand{\arraystretch}{1}
        \begin{tabular}{l *{4}{>{\centering\arraybackslash}p{1.1cm}}} 
            \toprule
             & \multicolumn{2}{c}{\textbf{NL \cmark}} & \multicolumn{2}{c}{\textbf{NL \xmark}} \\
            \cmidrule(lr){2-3} \cmidrule(lr){4-5}
            \multicolumn{1}{c}{\textbf{Model}} & \textbf{ST \xmark} & \textbf{ST \cmark} & \textbf{ST \xmark} & \textbf{ST \cmark} \\
            \midrule
            \multicolumn{5}{c}{\textit{\textbf{Closed-Source LLMs}}} \\
            \midrule
            Claude-3.7-Sonnet & 2.6 & 2.6 & 2.4 & 2.6 \\
            Claude-4-Sonnet & 3.2 & 4.5 & 3.2 & 3.9 \\
            Claude-Sonnet-4.5 & 3.9 & 4.5 & 3.0 & 3.6 \\
            GPT-4o-mini & 1.3 & 1.5 & 1.3 & 1.5 \\
            GPT-5 & 3.4 & 3.4 & 1.9 & 2.8 \\
            GPT-5-mini & 3.6 & 4.1 & 2.1 & 2.6 \\
            GPT-5-nano & 2.1 & 1.5 & 2.1 & 1.5 \\
            \midrule
            \multicolumn{5}{c}{\textit{\textbf{Open-Source LLMs}}} \\
            \midrule
            DeepSeek-R1 & 3.0 & 4.1 & 2.8 & 3.0 \\
            GPT-OSS-120B & 3.9 & 4.5 & 4.7 & 5.1 \\
            Llama-3.1-405B & 2.6 & 2.6 & 2.4 & 2.8 \\
            Llama-3.1-70B & 2.4 & 2.4 & 1.9 & 1.7 \\
            Llama-3.1-8B & 1.5 & 1.3 & 1.5 & 1.5 \\
            \midrule
            \multicolumn{5}{c}{\textit{\textbf{Math Specialized Models}}} \\
            \midrule
            DeepSeek-Prover-v2-7B & 2.4 & 3.2 & 2.8 & 3.2 \\
            Goedel-Formalizer-v2-8B & 1.5 & 2.4 & 1.9 & 2.1 \\
            Goedel-Prover-v2-32B & 2.1 & 3.2 & 1.3 & 0.9 \\
            Goedel-Prover-v2-8B & 1.7 & 2.6 & 2.4 & 3.0 \\
            Kimina-Autoformalizer-7B & 1.3 & 1.5 & 1.3 & 1.5 \\
            Kimina-Prover-Distill-8B & 2.6 & 1.7 & 1.7 & 2.6 \\
            \bottomrule
        \end{tabular}%
    }
    \caption{
    Impact of input context on TA (\%) on \ours Hard. 
    We compare two axes: whether an \gls{nl} proof is provided, and whether a successor theorem is provided (\cmark) or withheld (\xmark).
    }
    \label{tab:ablation_research_hard}
\end{table}

\Cref{tab:ablation_research_hard} presents the full ablation results on the \ours Hard subsets, varying whether a \gls{nl} proof and/or successor theorem is provided as context.

% On the full \ours benchmark (\Cref{tab:ablation_research}), the effect of successor theorem is less pronounced than on the Hard subset, with most models showing marginal differences across context settings. 
% This is expected, as the full benchmark includes simpler, library-heavy problems where compilation alone often suffices.

On \ours Hard (\Cref{tab:ablation_research_hard}), providing successor theorems yields consistent improvements for most models. 
GPT-OSS-120B achieves the highest \gls{ea} of 5.1\% in the NL \xmark, ST \cmark setting. 
Notably, \gls{nl} proofs show mixed or negligible effects: for several models (e.g., Claude-4-Sonnet, GPT-5-mini), adding \gls{nl} proofs without a successor theorem does not improve or even decreases \gls{ea} compared to the baseline (NL \xmark, successor theorem \xmark), while adding a successor theorem consistently helps regardless of whether \gls{nl} is provided.

\section{Compilation vs.\ Testing Accuracy}
\label{sec:appendix:compile_vs_ea}

\Cref{fig:model_size_scaling} visualizes the divergence between compilation accuracy and \gls{ea} on the \ours Hard subset. 
While compilation rates vary widely across models (4.3\% to 75.6\%) and generally improve with scale, \gls{ea} remains consistently below 5\% for nearly all models, visually confirming the persistent semantic gap discussed in \Cref{subsec:metric_comparison}.

\section{Detailed Experimental Results}
\label{sec:appendix:detailed_results}

\begin{table*}[t]
    \centering
    \scalebox{0.8}{
        \renewcommand{\arraystretch}{1.1}
        \begin{tabular}{lrrrrrr}
            \toprule
                                     & \multicolumn{2}{c}{\textbf{Existing}}                                  & \multicolumn{2}{c}{\textbf{\ours}}                                     & \multicolumn{2}{c}{\textbf{\ours (Hard)}} \\
            \cmidrule(lr){2-3} \cmidrule(lr){4-5} \cmidrule(lr){6-7}
            \textbf{Model}           & \textbf{Compile}                      & \textbf{\gls{ea}}              & \textbf{Compile} & \textbf{\gls{ea}} & \textbf{Compile} & \textbf{\gls{ea}} \\
            \midrule
            \multicolumn{7}{c}{\textit{\textbf{Closed-Source LLMs}}} \\
            \midrule
            Claude-3.7-Sonnet        & 79.5 & 76.7 & 75.1 & 36.8 & 34.3 & 2.6 \\
            Claude-4-Sonnet          & 86.6 & 83.3 & 81.1 & 36.0 & 47.5 & 4.5 \\
            Claude-Sonnet-4.5        & 90.6 & 88.1 & 80.3 & 38.9 & 46.0 & 4.5 \\
            GPT-4o-mini              & 64.3 & 60.5 & 59.1 & 36.7 & 12.2 & 1.5 \\
            GPT-5                    & 92.4 & 80.5 & 85.7 & 37.7 & 68.3 & 3.4 \\
            GPT-5-mini               & 86.3 & 80.8 & 86.0 & 37.9 & 66.6 & 4.1 \\
            GPT-5-nano               & 89.6 & 69.6 & 88.7 & 36.6 & 75.6 & 1.5 \\
            \midrule
            \multicolumn{7}{c}{\textit{\textbf{Open-Source LLMs}}} \\
            \midrule
            DeepSeek-R1              & 82.3 & 79.0 & 70.5 & 36.3 & 25.1 & 4.1 \\
            GPT-OSS-120B             & 53.7 & 48.4 & 68.2 & 32.3 & 40.0 & 4.5 \\
            Llama-3.1-405B           & 75.4 & 72.9 & 63.4 & 33.2 & 27.8 & 2.6 \\
            Llama-3.1-70B            & 72.2 & 68.1 & 65.4 & 37.0 & 20.1 & 2.4 \\
            Llama-3.1-8B             & 54.4 & 45.8 & 36.9 & 24.8 & 6.4 & 1.3 \\
            \midrule
            \multicolumn{7}{c}{\textit{\textbf{Math Specialized Models}}} \\
            \midrule
            DeepSeek-Prover-v2-7B    & 63.8 & 53.9 & 62.2 & 30.0 & 35.5 & 3.2 \\
            Goedel-Formalizer-v2-8B  & 22.0 & 19.2 & 38.7 & 23.6 & 22.1 & 2.4 \\
            Goedel-Prover-v2-32B     & 61.3 & 54.2 & 74.6 & 31.0 & 54.6 & 3.2 \\
            Goedel-Prover-v2-8B      & 22.3 & 18.7 & 46.1 & 24.5 & 37.9 & 2.6 \\
            Kimina-Autoformalizer-7B & 14.2 & 1.0 & 21.9 & 20.0 & 4.3 & 1.5 \\
            Kimina-Prover-Distill-8B & 3.0 & 2.0 & 28.0 & 23.4 & 5.1 & 1.7 \\
            \bottomrule
        \end{tabular}
    }
    \caption{
    Full results across three evaluation settings: existing benchmarks, \ours (full), and \ours Hard. 
    Results are grouped by model type.
    }
    \label{tab:main_results_difficulty}
\end{table*}

\Cref{tab:main_results_difficulty} presents the full compilation and testing accuracy across three evaluation settings: existing benchmarks, \ours (full), and \ours Hard.

On existing benchmarks, the gap between compilation and testing accuracy is relatively small for most models. 
For instance, Claude-Sonnet-4.5 achieves 90.6\% compilation and 88.1\% testing accuracy.
This is expected, as these benchmarks consist of isolated problems with a small number of successor theorems, limiting the discriminative power of \gls{ea}.

On the full \ours benchmark, a clear gap emerges. Models that compile at 60-80\% drop to 24-39\% testing accuracy, reflecting the stricter verification imposed by successor theorems in repository-level problems.
Notably, testing accuracy across models converges to a narrow range (24-39\%) despite wide variation in compilation rates, suggesting that most models can produce syntactically valid code for repository-level problems but struggle equally with semantic correctness.

On \ours Hard, the gap becomes most pronounced.
Compilation rates vary widely (4.3\% to 75.6\%), yet testing accuracy is uniformly low (1.3\%-4.5\%).
GPT-5-nano achieves the highest compilation accuracy (75.6\%) but only 1.5\% \gls{ea}, a 50$\times$ gap, while Claude-Sonnet-4.5 compiles at 46.0\% but achieves the highest \gls{ea} (4.5\%).
This rank reversal further demonstrates that compilation accuracy is a poor predictor of semantic correctness and that the Hard subset effectively isolates the challenge of genuine mathematical reasoning.

% \clearpage
% \onecolumn

\section{Discussion}
We now examine the scope and reliability of \gls{ea} in light of the limitations above.

While coverage imposes an upper bound on what \gls{ea} can detect, the risk of false negatives, where a logically correct but differently formulated theorem fails to compile against successor theorems, is minimal in our setup.
The type declarations such as \texttt{theorem}, \texttt{lemma}, and \texttt{def}, and the exact name of the target are provided as context (\Cref{subsec:integration}, footnote 1), and successor theorems reference the target by this name.
If a generated theorem is logically correct, it should type-check successfully against all successor theorems, leaving little room for false negatives.

The inapplicability of \gls{ea} to standalone theorems is likewise bounded in practice: across our collected repositories, approximately 1.4\% of theorems have no successor theorems.
Existing methods, such as BLEU or manual inspection, can be applied to these few cases, while our contribution enables automatic semantic evaluation for the remaining 98.6\%.

\section{Future Work}
Our results suggest several promising directions for extending \gls{ea} and \ours.
A natural next step is to refine the binary scoring of \gls{ea} into a graded metric that accounts for partial success in successor theorems compilation, providing a more nuanced signal for model development.
Beyond Lean~4, \gls{ea} is language-agnostic by design and can be extended to any proof assistant that provides a dependency parser and research-level repositories, such as Coq and Isabelle.\footnote{Coq provides \texttt{coq-dpdgraph} and the built-in \texttt{coqdep} for dependency parsing, with more than 45 research-level packages on GitHub. Isabelle provides a built-in session dependency system, and the Archive of Formal Proofs contains more than 935 entries.} We chose Lean 4 for its actively growing ecosystem with diverse repositories, and leave extension to other proof assistants for future work.
Within Lean~4 itself, expanding \ours to additional repositories as the ecosystem grows would further improve domain coverage.
We also plan to pursue a larger-scale verification of the natural language annotations in consultation with Lean mathematics experts, complementing our initial sample-based assessment.

\section{Prompts}
\label{sec:appendix:prompts}

Following standard conventions in code generation evaluation, we design structured prompts that provide clear task specifications and output format constraints.
Similar to how SWE-Bench restricts models to bash-only interactions with explicit formatting requirements, our prompts enforce a strict output protocol: models must produce exactly one Lean 4 code block preceded by a reasoning section.
The full prompt templates for informalization and autoformalization are provided in \Cref{fig:prompt-informalize} and \Cref{fig:prompt-autoformalize}, respectively.
We do not perform prompt sensitivity analysis (e.g., varying instruction phrasing or format constraints), as our goal is to benchmark model capabilities under a standardized evaluation protocol rather than optimize for prompt engineering.
This is consistent with prior work in both code generation (HumanEval, APPS) and theorem proving (MiniF2F), where a fixed prompt template is used across all models to ensure fair comparison.

\clearpage
\onecolumn
% System Prompt
\begin{systemprompt}
    You are a helpful assistant that can interact multiple times with a computer shell to solve programming tasks.
    However, for this specific interaction, you are acting as an expert Lean 4 mathematician tasked with translating formal Lean 4 code into a natural language description or proof.

    Your response must contain exactly ONE plaintext code block.

    Include a THOUGHT section before your code where you explain your reasoning process.
    Format your response as shown in \texttt{<format\_example>}.

    \texttt{<format\_example>}
    \begin{textcode}
THOUGHT: Your reasoning and analysis here
    \end{textcode}

    \begin{textcode}
your_natural_language_description_here
    \end{textcode}
    \texttt{</format\_example>}

    Failure to follow these rules will cause your response to be rejected.
\end{systemprompt}

% User Prompt
\begin{userprompt}
    \texttt{<task\_description>}
    \\
    Your task is to provide a natural language description (that is informal proof or problem statement) for the specific Lean 4 declaration.
    \\
    Target Code Name: \texttt{\{\{ target\_code\_name \}\}}
    \\
    The declaration is part of a larger code sequence.

    The Code Context is provided below.
    \begin{itemize}
        \item \texttt{Header}: Imports and open namespaces.
        \item \texttt{Code Before}: Code appearing before the target.
        \item \texttt{Target Code}: The specific code you need to describe/informalize.
    \end{itemize}

    Header:
    \begin{leancode}
{{ header }}
    \end{leancode}

    Code Before:
    \begin{leancode}
{{ before_target_code }}
    \end{leancode}

    Target Code:
    \begin{leancode}
{{ target_code }}
    \end{leancode}

    Code After:
    \begin{leancode}
{{ after_target_code }}
    \end{leancode}

    \texttt{<instructions>}
    \\
    \textbf{\# Task Instructions}

    \textbf{\#\# Overview}
    \\
    You are an expert Lean 4 mathematician.
    You must provide a clear, concise, and mathematically accurate natural language description of the \texttt{Target Code}.
    \begin{itemize}
        \item If \texttt{Target Code} is a theorem/proof, provide an informal proof.
        \item If \texttt{Target Code} is a definition, provide a mathematical definition in latex.
        \item Use standard mathematical terminology (TeX/LaTeX style for formulas is preferred).
        \item You must not use the exact function name used in lean code. You need to describe in natural language.
        \item Do not start with which function name you are translating. Just start with the informal proof or problem statement.
    \end{itemize}

    \textbf{\#\# Output Coding Rules}
    \\
    1. A \textbf{THOUGHT} section where you explain your reasoning:
    \begin{itemize}
        \item Analyze the \texttt{Target Code} to understand what it proves or defines.
        \item Consider the context from \texttt{Code Before} and \texttt{Code After} if relevant.
        \item Plan the natural language description.
    \end{itemize}
    2. A single plaintext code block with your description.

    Format your responses like this:

    \texttt{<format\_example>}
    \\
    \begin{textcode}
THOUGHT: The target code defines a function \texttt{foo} that adds two numbers.
    I will describe it as "The function determines the sum of two integers."
    \end{textcode}

    \begin{textcode}
The function determines the sum of two integers.
    \end{textcode}
    \texttt{</format\_example>}

    \textbf{CRITICAL REQUIREMENTS:}
    \begin{itemize}
        \item Your response SHOULD include a THOUGHT section.
        \item Your response MUST include EXACTLY ONE plaintext block.
        \item This block MUST contain the COMPLETELY NEW natural language description.
        \item Do NOT repeat the input code.
    \end{itemize}
    \texttt{</instructions>}
\end{userprompt}

% AI Response
\begin{modelresponse}
    \begin{textcode}
{{ thoughts }}
    \end{textcode}

    \begin{textcode}
{{ target_natural_language }}
    \end{textcode}
\end{modelresponse}

\captionof{figure}{Prompt template for informalizing Lean 4 code.}
\label{fig:prompt-informalize}

% System Prompt
\begin{systemprompt}
    You are a helpful assistant that can interact multiple times with a computer shell to solve programming tasks.
    However, for this specific interaction, you are acting as an expert Lean 4 mathematician tasked with implementing a formal Lean 4 declaration (theorem or definition) based on the provided context.

    Your response must contain exactly ONE lean code block.

    Include a THOUGHT section before your code where you explain your reasoning process.
    Format your response as shown in \texttt{<format\_example>}.

    \texttt{<format\_example>}
    \begin{textcode}
THOUGHT: Your reasoning and analysis here, explaining how you deduce the implementation from the context.
    \end{textcode}

    \begin{leancode}
your_formal_lean_code_here
    \end{leancode}
    \texttt{</format\_example>}

    Failure to follow these rules will cause your response to be rejected.
\end{systemprompt}

% User Prompt
\begin{userprompt}
    \texttt{<task\_description>}
    \\
    Your task is to provide the full formal Lean 4 code for the specific declaration named \texttt{\{\{ target\_code\_name \}\}}.
    The declaration is part of a larger code sequence, and you must implement it to fit seamlessly into the provided context.

    The Code Context is provided below.
    \begin{itemize}
        \item \texttt{Header}: Imports and open namespaces.
        \item \texttt{Code Before}: Code appearing before the target.
        \item \texttt{Target Code Name}: The name of the definition or theorem you need to implement.
        \item \texttt{Natural Language Description}: A description of the target code in natural language.
        \item \texttt{Code After}: Code appearing after the target.
    \end{itemize}

    Header:
    \begin{leancode}
{{ header }}
    \end{leancode}

    Code Before:
    \begin{leancode}
{{ before_target_code }}
    \end{leancode}

    Target Code Name: 
    \begin{textcode}
{{target_code_name}}
    \end{textcode}

    Natural Language Description:
    \begin{textcode}
{{target_natural_language}}
    \end{textcode}

    Code After:
    \begin{leancode}
{{ after_target_code }}
    \end{leancode}

    \texttt{<instructions>}
    \\
    \textbf{\# Task Instructions}
    \\
    \textbf{\#\# Overview}
    \\
    You are an expert Lean 4 mathematician.
    You must provide the complete and correct Lean 4 implementation for \texttt{\{\{ target\_code\_name \}\}}.
    \begin{itemize}
        \item If it is a theorem, provide the statement and the full proof (do not use \texttt{sorry}).
        \item If it is a definition, provide the full definition.
        \item Pay close attention to \texttt{Code Before} and \texttt{Code After} to infer the correct type signatures, variable names, and logical dependencies.
        \item Ensure your code compiles and integrates correctly with the surrounding context.
    \end{itemize}

    \textbf{\#\# Output Coding Rules}
    \\
    1. A \textbf{THOUGHT} section where you explain your reasoning:
    \begin{itemize}
        \item Analyze the \texttt{Code Before} and \texttt{Code After} to deduce the purpose and signature of \texttt{\{\{ target\_code\_name \}\}}.
        \item Plan the implementation.
    \end{itemize}
    2. A single lean code block with your implementation.

    Format your responses like this:

    \texttt{<format\_example>}
    \\
    \begin{textcode}
THOUGHT: The context shows a function \texttt{foo} is used to add two numbers.
I will implement \texttt{foo} as a definition taking two Nats and returning their sum.
    \end{textcode}

    \begin{leancode}
def foo (n m : Nat) : Nat := n + m
    \end{leancode}
    \texttt{</format\_example>}

    \textbf{CRITICAL REQUIREMENTS:}
    \begin{itemize}
        \item Your response SHOULD include a THOUGHT section.
        \item Your response MUST include EXACTLY ONE lean block.
        \item This block MUST contain the COMPLETELY NEW implementation for \texttt{\{\{ target\_code\_name \}\}}.
        \item Do NOT repeat the \texttt{Header} or \texttt{Code Before} or \texttt{Code After}. Only output the code for \texttt{\{\{ target\_code\_name \}\}}.
    \end{itemize}
    \texttt{</instructions>}
\end{userprompt}

% AI Response
\begin{modelresponse}
    \begin{textcode}
{{thoughts}}
    \end{textcode}

    \begin{leancode}
{{ formal_proof }}
    \end{leancode}
\end{modelresponse}

\captionof{figure}{Prompt template for autoformalizing to Lean 4 code.}
\label{fig:prompt-autoformalize}

\end{document}